\documentclass[conference]{IEEEtran}

\usepackage{tocloft}
\usepackage{algorithm}
\usepackage{algpseudocode}
\usepackage{amsmath,amssymb,amsfonts,amsbsy, amssymb}
\usepackage{textcomp}
\usepackage{bm}
\DeclareMathOperator*{\argmax}{argmax}
\usepackage[table]{xcolor} 
\usepackage{graphicx}
\usepackage{stfloats}
\usepackage{pdfpages}
\usepackage{makecell}
\usepackage{color,soul}

\usepackage{geometry}
\usepackage{datetime2}
\usepackage{url}
\usepackage{array}
\usepackage{booktabs}
\usepackage{multirow}
\usepackage{tabularx}
\usepackage{threeparttable}
\usepackage[normalem]{ulem}
\usepackage{ragged2e}
\usepackage{enumitem}
\usepackage{caption}
\usepackage{subcaption}

\ifCLASSOPTIONcompsoc
  \usepackage[nocompress]{cite}
\else
  \usepackage{cite}
\fi

\usepackage{listings}

\usepackage{float}
\usepackage{placeins}

\usepackage[pagebackref=false,breaklinks=true,colorlinks=true,bookmarks=false]{hyperref}

\definecolor{robo_blue}{RGB}{66,133,244}
\definecolor{robo_red}{RGB}{231,66,52}
\definecolor{robo_yellow}{RGB}{251,189,5}
\definecolor{robo_green}{RGB}{51,168,82}
\definecolor{robo_gray}{RGB}{165,165,165}
\definecolor{darkgreen}{rgb}{0.0,0.5,0.0}
\definecolor{treecomment}{RGB}{150,150,150}
\definecolor{pc_keyword}{RGB}{0,128,128}
\definecolor{pc_comment}{RGB}{0,128,0}
\definecolor{pc_number}{RGB}{255,140,0}
\definecolor{pc_operator}{RGB}{128,0,128}
\definecolor{pc_string}{RGB}{200,0,0}

\lstdefinestyle{caibase}{
  backgroundcolor=\color{gray!10},
  basicstyle=\scriptsize\ttfamily,
  frame=single,
  rulecolor=\color{black!30},
  breaklines=true,
  columns=fullflexible,
  captionpos=b,
  numbersep=8pt
}

\lstdefinestyle{pseudocode}{
    style=caibase,
    numbers=left,
    numberstyle=\tiny\color{robo_gray},
    keywordstyle=\color{pc_keyword}\bfseries,
    commentstyle=\color{pc_comment}\itshape,
    stringstyle=\color{pc_string},
    morekeywords={IF,ELSE,ENDIF,STORE,CALL,CREATE,BEGIN,END,FOR,WHILE,RETURN,Input,Output,ENDWHILE,START,THEN,UPDATE,DISPLAY,INITIALIZE,CAPTURE,ALLOW,SEND,CLICK,RECEIVE,WAIT,ENDFOR,DO},
    sensitive=true,
    literate=
        {→}{{\textcolor{pc_operator}{$\rightarrow$}}}1
        {<-}{{\textcolor{pc_operator}{$\leftarrow$}}}1
        {=}{{\textcolor{pc_operator}{=}}}1
        {/}{{\textcolor{pc_operator}{/}}}1
        {:}{{\textcolor{pc_operator}{:}}}1
        {+}{{\textcolor{pc_operator}{+}}}1,
    tabsize=2
}

\lstdefinestyle{directory}{
  style=caibase,
  comment=[l]{\#},
  commentstyle=\color{treecomment}\itshape
}

\lstdefinestyle{json}{
  style=caibase,
  frame=single,
  breaklines=true,
  commentstyle=\color{treecomment}\itshape,
  stringstyle=\color{pc_string},
  showstringspaces=false,
  moredelim=[s][\color{pc_string}]{"}{"},
  literate=
   *{true}{{{\color{robo_green}true}}}{4}
    {false}{{{\color{robo_red}false}}}{5}
    {null}{{{\color{robo_gray}null}}}{4}
}

\usepackage[explicit]{titlesec}
\titleformat{\section}
  {\normalfont\fontsize{11}{12}\bfseries\centering}{\thesection}{1em}{#1}
\titleformat{\subsection}[block]
  {\normalfont\normalsize\bfseries}{\thesubsection}{1em}{#1}
\titleformat{\subsubsection}[block]
  {\normalfont\normalsize\bfseries}{\thesubsubsection}{1em}{#1}

\begin{document}

    
    


    
    
    
    
\setcounter{page}{1}

\title{HumP-KD: A Hybrid Uncertainty-Aware Multi-Stage Progressive Knowledge Distillation Framework for Efficient Fire Classification}

\author{
\IEEEauthorblockN{Mohammed Arif Mainuddin}
\IEEEauthorblockA{\textit{Electrical and Computer Engineering} \\
\textit{North South University}\\
Dhaka, Bangladesh \\
\href{mailto:mohammed.mainuddin@northsouth.edu}{mohammed.mainuddin@northsouth.edu}}
\and
\IEEEauthorblockN{Najifa Tabassum}
\IEEEauthorblockA{\textit{Electrical and Computer Engineering} \\
\textit{North South University}\\
Dhaka, Bangladesh \\
\href{mailto:najifa.tabassum@northsouth.edu}{najifa.tabassum@northsouth.edu}}
\and
\IEEEauthorblockN{Omar Ibne Shahid}
\IEEEauthorblockA{\textit{Electrical and Computer Engineering} \\
\textit{North South University}\\
Dhaka, Bangladesh \\
\href{mailto:omar.shahid@northsouth.edu}{omar.shahid@northsouth.edu}}
\and
\IEEEauthorblockN{Riasat Khan}
\IEEEauthorblockA{\textit{Electrical and Computer Engineering} \\
\textit{North South University}\\
Dhaka, Bangladesh \\
\href{mailto:riasat.khan@northsouth.edu}{riasat.khan@northsouth.edu}}

}

\maketitle
\thispagestyle{plain}
\pagestyle{plain}

\IEEEdisplaynontitleabstractindextext
\ifCLASSOPTIONpeerreview
\begin{center} \bfseries EDICS Category: 3-BBND \end{center}
\fi
\IEEEpeerreviewmaketitle

\begin{abstract}
Fire accidents have become a growing concern for both humans and wildlife, while also threatening the climate by contributing to global warming. Real-time fire classification systems require models that are simultaneously accurate, computationally efficient, and deployable on resource-constrained hardware. This work proposes \textbf{HumP-KD}, a Hybrid Uncertainty-aware Multi-stage Progressive Knowledge Distillation framework for efficient fire classification. Two datasets, FlameVision and Dataset-II, containing 8,600 and 31,309 images, are used. Various CNN and transformer baselines are applied under standard preprocessing, online augmentation, Gaussian noise and motion blur robustness conditions. The proposed HumP-KD model distills knowledge from two frozen heterogeneous transformer teachers, Swin-Tiny and ViT-Base, along with their Meta-MLP ensemble, into a lightweight MobileViT-S student via three tightly integrated components. Uncertainty-Aware Knowledge Distillation dynamically weights teacher contributions based on inter-teacher prediction variance. Hierarchical Progressive Knowledge Distillation employs a Hierarchical Feature Builder. It generates a fused spatial attention mask to guide distillation toward discriminative regions selectively. Multi-Stage Knowledge Distillation progressively activates three distillation stages across training. On Dataset-II, HumP-KD achieves a mean F1 score of $0.9876 \pm 0.0063$ across 10 independent trials, significantly outperforming the MobileViT-S baseline trained without distillation ($0.9537 \pm 0.0351$), with statistical significance confirmed by both independent t-test ($p = 0.0195$) and Wilcoxon signed-rank test ($W = 1$, $p = 0.0039$). The proposed method also demonstrates strong generalization across datasets and robustness under degraded visual conditions. The student model retains only 4.94M parameters and 19.01Mb model size, representing a $5.7\times$ parameter reduction over Swin-Tiny and a $17.5\times$ reduction over ViT-Base, while achieving 37.72 CPU FPS, making it suitable for real-time deployment. The distilled student model is deployed as a web-based fire classification application, returning a confidence score and a Grad-CAM heatmap for each input image.
\end{abstract}


\section{Introduction}

Fire is one of the most destructive phenomena caused by both natural and human-introduced hazards. These fire-causing scenarios have consistently been recognized as a noxious hazard in urban and wildlife environments, causing ecosystem destruction, loss of life, financial damage, and many other threats ~\cite{zia20243enb2}. Fire incidents can be natural or man-made; for instance, fire can be caused by natural events such as lightning strikes, excessive heat, dry vegetation, or by human carelessness, such as electrical faults, arson, and unattended stoves~\cite{zia20243enb2}. 

Fire causes damage to properties and lives~\cite{sathishkumar2023forest}. In
2024, USA witnessed 1.39 million fire incidents, resulting in US\$19 billion in losses, with 4000 deaths. Densely populated areas, shopping malls, restaurants, and garment factories are the red zones for massive fire outbreaks. According to the fire service, common causes of major fire outbreaks include gas cylinder leakage, unsafe wiring, poor maintenance, overloaded connections, and burning cigarette butts~\cite{dailyStarFire2025}.
Before the advancement of AI, fire detection relied on manual surveillance systems, such as CCTV with smoke and thermal sensors, observation towers, and safety alarm systems~\cite{zia20243enb2}. To reduce fire-causing deaths, modern fire detection techniques have become necessary~\cite{yang2024real}. These automatic techniques enhance accuracy by learning to handle complex data with different preprocessing methods.

This work introduces \textbf{HumP-KD}, a novel hybrid knowledge distillation framework for efficient fire classification. The key contributions are summarized as follows:

\begin{itemize}

\item \textbf{HumP-KD Framework:} A novel Hybrid Uncertainty-aware Multi-stage Progressive Knowledge Distillation framework is proposed, consisting of three tightly integrated components: (i) Uncertainty-Aware Knowledge Distillation (UAKD), which dynamically weights teacher contributions based on inter-teacher prediction variance; (ii) Hierarchical Progressive Knowledge Distillation (HPKD), which employs a Hierarchical Feature Builder (HFBuilder) to generate a fused spatial attention mask guiding distillation toward discriminative regions; and (iii) Multi-Stage Knowledge Distillation (MSKD), which progressively activates three distillation stages across training. 

\item \textbf{Statistical Significance Analysis:} The performance improvement of HumP-KD over the MobileViT-S baseline trained without distillation is validated through 10 independent trials, with statistical significance confirmed by both an independent t-test and Wilcoxon signed-rank test.

\item \textbf{Robustness Evaluation:} HumP-KD is evaluated under corrupted image conditions, including Gaussian noise and motion blur on both datasets, demonstrating its stability under real-world degraded visual conditions. 

\item \textbf{Cross-Dataset Generalization:} A cross-evaluation is conducted by training on one dataset and testing on the other, validating the generalization capability of HumP-KD across different fire scenarios and imaging conditions. 
\item \textbf{Qualitative Attention Analysis:} 
Multi-scale attention rollout visualizations are presented for the proposed model, demonstrating progressive attention localization toward discriminative fire regions across distillation stages. 

\item \textbf{Real-Time Web Deployment:} The distilled student model is deployed as a web-based fire classification application, returning a confidence score and a Grad-CAM heatmap for each input image.

\end{itemize}

The novelty of this work lies in a unified hybrid knowledge distillation framework that integrates uncertainty-aware weighting, hierarchical attention-guided learning, and multi-stage progressive training to enable lightweight models for efficient fire classification.


\section{Related Works}

Traditional fire detection typically involves the use of the watchtower technique and smoke, heat, fume and various detectors. These conventional techniques' slow response, false alarms, maintenance, and repair costs are negative factors that hinder efficient utilization. As a result, advanced AI-based automatic fire detection has been introduced in recent studies. This section discusses recent deep learning-based works for classifying fire incidents.

\subsection{Deep Learning and Hybrid Architectures for Fire Detection}

Veerappampalayam et al. \cite{sathishkumar2023forest} applied transfer learning without forgetting (LWF) to preserve the network's prior capabilities. LWF with the Xception approach improved the accuracy from 79.23\% to 91.41\% for the BowFire dataset. Yang et al. \cite{yang2024real} implemented a cloud-edge collaborative system for fire and smoke detection employing an iterative transfer learning technique. The applied augmentation and coordinated attention for global feature extraction achieved 96.4\% accuracy. Zia et al. \cite{zia20243enb2} introduced an end-to-end neural network model, 3ENB2 (based on EfficientNetB2), to detect fires from images. The proposed model with online data augmentation achieved 99.04\% accuracy. Choi et al. \cite{10839387} proposed the early detection of wildfire and no-fire using vision transformer-based models. The authors employ Swin Transformers as the backbone of Mask-RCNN, which helps detect bounding boxes and segmentation masks. Compared to the ResNet-50 model, the Swin Transformer-based model achieved a higher Segm mAP50 of 0.842. Ahmad et al. \cite{fire8060211} applied the CN2VF-Net model, which combines CNN and ViT. The combination learns a multiscale attention mechanism that dynamically focuses on fire regions with different scales. The authors applied the CNN-ViT model to the D-Fire dataset, achieving an F1 score of 81.5\% and a precision of 83.3\%. Madafri et al. \cite{el2024real} proposed multi-task knowledge distillation, trained on a hierarchically structured wildfire dataset. The applied approach with a combined loss function achieved 93.36\% accuracy and a 97.51\% F1 score. Ahmed et al. \cite{ahmad2025firenet} introduced FireNet-KD, which ensembles a Vision Transformer with a CNN as a parent model. The parent model extracts hierarchical features and spatial locality, then distills them into a Swin Transformer-based student model. The FireNet-KD technique achieves 95.16\% accuracy and 97.31\% mAP@50. Li et al. \cite{11071024} developed a Multiscale Spatial Attention mechanism (MSA) to detect fires. The authors applied the dual distillation technique by designing two distinct modules to ensure robust feature extraction and cross-talk consistency. 

\subsection{Knowledge Distillation Strategies}

Zeng et al. \cite{bioengineering11010070} implemented a dual-stage progressive knowledge distillation technique for skin disease classification. The dual-stage knowledge distillation technique consists of several fusion steps. The feature-based distillation is integrated and hierarchical fusion is applied with a masking strategy. Ousalah et al.~\cite{ousalah2025uncertainty} proposed uncertainty-aware knowledge distillation, in which they quantified the teacher prediction using inter-model variance and dynamically weighted it into the distillation loss. This approach ensures the student learns more from reliable teacher predictions and less from uncertain ones. Pavel et al.~\cite{pavel2025multi} introduced multi-stage knowledge distillation, in which they progressively transferred knowledge across hierarchical teacher layers at different training stages.

\subsection{Lightweight and UAV-based Fire Detection Models}

Zhai et al. \cite{11128176} developed an aerial fire-detection model that achieved higher detection accuracy and ease of deployment. The LAFNET model integrates a lightweight block with the path area network. The LAFNET model improved mAP by 2.1\% while keeping the model lightweight by reducing the number of parameters by 27.8\% on the FlameVision dataset. Gao et al. \cite{Gao2025YOLO11RLNAA} applied an UAV-based algorithm to detect forest fires. The authors replaced the YOLO11 backbone with the RepVGG technique for improved feature extraction, fireline texture fusion attention and nano-optimization. The YOLO11-RLN technique outperforms YOLO11, achieving 7.338\% improvement in precision and 5.392\% in recall. Li et al. \cite{10.1145/3744103.3744121} introduced the Fire Identification and Classification Network (FICNet) to reduce false positive detections and improve fire classification and detection accuracy. The FICNet achieves a precision of 0.784 on the FlameVision aerial image dataset. Titu et al. \cite{titu2024real} used advanced deep learning models to detect fires in real time with a Raspberry Pi 5 and a drone. The authors used a combined dataset and implemented Knowledge Distillation using DETR techniques. This approach achieved 95.21\% detection accuracy and an F1 score of 0.985. On the edge device, it achieved 89.23\% accuracy and a frame rate of 8 for the live experiment.

Researchers efficiently implemented various approaches, such as transfer learning, ensemble learning with parent models, learning without forgetting, and others, for fire detection. All existing work that focuses on increasing accuracy overlooks the development of a lightweight model. To our knowledge, departing from the conventional vanilla knowledge distillation approach, uncertainty-aware~\cite{ousalah2025uncertainty} and multi-stage~\cite{pavel2025multi} distillation strategies have been explored independently. No prior work has unified these approaches with hierarchical attention masking into a single hybrid framework for fire classification. HumP-KD addresses this gap by integrating UAKD, HPKD, and MSKD into a cohesive distillation pipeline for fire classification.

\section{Proposed System}

\subsection{Problem Formulation}

The proposed system classifies fire and no-fire using the FlameVision dataset and a unified hybrid knowledge distillation framework. Let the input dataset $\boldsymbol{\mathcal{D}}$ consist of fire and no-fire images $\boldsymbol{x_i} \in \mathbb{R}^{H \times W \times 3}$, where $\boldsymbol{H}$ and $\boldsymbol{W}$ represent height and width, and $3$ denotes the RGB channels. The binary label set is $\boldsymbol{\mathcal{Y}} = \{0, 1\}$, where $0$ denotes fire and $1$ denotes no-fire. The goal is to train a lightweight student model $\boldsymbol{f_s}$ that produces a logit vector $\boldsymbol{\hat{y}} = f_s(x) \in \mathbb{R}^C$, and predicted class $\boldsymbol{\hat{y}_C} = \argmax(\hat{y})$, where $C = 2$ is the number of classes.

A heterogeneous dual teacher ensemble $\boldsymbol{\mathcal{T}} = \{f_{\text{swin}}, f_{\text{vit}}\}$ is employed, consisting of a pre-trained Swin Transformer $\boldsymbol{f_{\text{swin}}}$ and a Vision Transformer $\boldsymbol{f_{\text{vit}}}$, both fine-tuned on the target domain with classification heads replaced to output $C = 2$ classes. A Meta-MLP $\boldsymbol{f_{\text{meta}}}$ combines teacher logits into a refined ensemble prediction $\boldsymbol{\hat{y}_{\text{meta}}} = f_{\text{meta}}(\hat{y}_{\text{swin}},\ \hat{y}_{\text{vit}})$ [shown in Figure~\ref{fig:ensemble_architecture}]. During student training, all teacher parameters remain frozen.

The student $\boldsymbol{f_s}$ is a MobileViT-small architecture with intermediate feature representations extracted at three hierarchical stages $\boldsymbol{\{\mathcal{Z}_s^0,\ \mathcal{Z}_s^1,\ \mathcal{Z}_s^2\}}$ via forward hooks at stages $\{2, 3, 4\}$. Correspondingly, teacher feature representations are extracted as $\boldsymbol{\{\mathcal{Z}_t^0,\ \mathcal{Z}_t^1,\ \mathcal{Z}_t^2\}}$ from Swin Layer$_0$, ViT Block$_6$, and Swin Layer$_2$, respectively.

\subsection{Dataset Details}
This work uses the FlameVision dataset~\cite{flamevision2023}, which contains images of two categories, i.e., fire and no-fire. It contains aerial wildfire images, comprising 8600 high-resolution photos, with 5000 fire images and 3600 no-fire images. Another dataset utilized in the last phase of HumP-KD is the merged version of two publicly available datasets: Wildfire Dataset~\cite{el2023wildfire} and Forest Fire Smoke Dataset~\cite{minha2023forestfire}, referred to as \textbf{Dataset-II}. The merged Dataset-II contains two classes, fire and no-fire, with 20,886 training images, 3,001 validation images, and 7,422 test images.


\begin{table}[ht]
\centering
\caption{Distribution of images across training, validation, and test splits for FlameVision and Dataset-II}
\label{tab:dataset_distribution}
\resizebox{\linewidth}{!}{%
\begin{tabular}{lrrrr}
\toprule
\multirow{2}{*}{\textbf{Split}} & \multicolumn{2}{c}{\textbf{FlameVision}} & \multicolumn{2}{c}{\textbf{Dataset-II}} \\
\cmidrule(lr){2-3} \cmidrule(lr){4-5}
 & \textbf{Fire} & \textbf{No-fire} & \textbf{Fire} & \textbf{No-fire} \\
\midrule
Train        & 3600 & 3200 & 10229 & 10657 \\
Validation   & 700  & 200  & 1455  & 1546  \\
Test         & 700  & 200  & 3671  & 3751  \\
\midrule
\textbf{Total} & \textbf{5000} & \textbf{3600} & \textbf{15355} & \textbf{15954} \\
\bottomrule
\end{tabular}%
}
\end{table}

Table~\ref{tab:dataset_distribution} shows the distribution of fire and no-fire images across different splits (train, validation, and test) in both FlameVision and Dataset-II datasets. Fine-tuning of HumP-KD was implemented in two steps using the Dataset-II: First, a random subset of 1,000 images from the training set and 700 images from the validation set was selected. Secondly, the student model is trained with a small number of epochs. Finally, the HumP-KD model is tested with the complete test set of Dataset-II.

\subsection{Dataset Preprocessing}

The FlameVision dataset was split into training, validation, and test sets in an 8:1:1 ratio. Standard preprocessing steps were applied prior to model training. The preprocessing stage includes data cleaning, resizing to 224$\times$224, and applying normalization to rescale the image range. To avoid increasing the dataset size, we perform online augmentation, where transformations are applied dynamically during training without storing augmented data. We added speckle noise, random brightness, and adjusted saturation as online augmentation techniques to make the models more robust.

\subsection{Applied Student and Teacher Models}

In this work, MobileViT serves as the student model. HPKD enables MobileViT to learn not only logits but also selective features from two heterogeneous teachers (Swin and ViT). Through feature mapping, MobileViT becomes aware of ViT's global context and Swin's local features. Learning from two heterogeneous teachers in a hybrid manner, MobileViT provides better and more generalized output.

\textbf{Vanilla Knowledge Distillation and Hybrid Progressive Knowledge Distillation.}
Knowledge distillation is the process of transferring the teacher (heavy) model's soft labels to the student (lightweight) model. The student then combines the teacher's soft labels with its own hard labels, learns the teacher model's patterns, and finally delivers the best evaluation. Unlike traditional vanilla KD, where the student model only learns from the teacher model's soft labels, the student model in HPKD learns from both the teacher's ignored and enriched areas, as well as the teacher's logits \cite{bioengineering11010070}. In HPKD, the student model learns from two heterogeneous teacher models in a hybrid progressive way.

\begin{figure*}[!t]
    \centering
    \includegraphics[width=\textwidth]{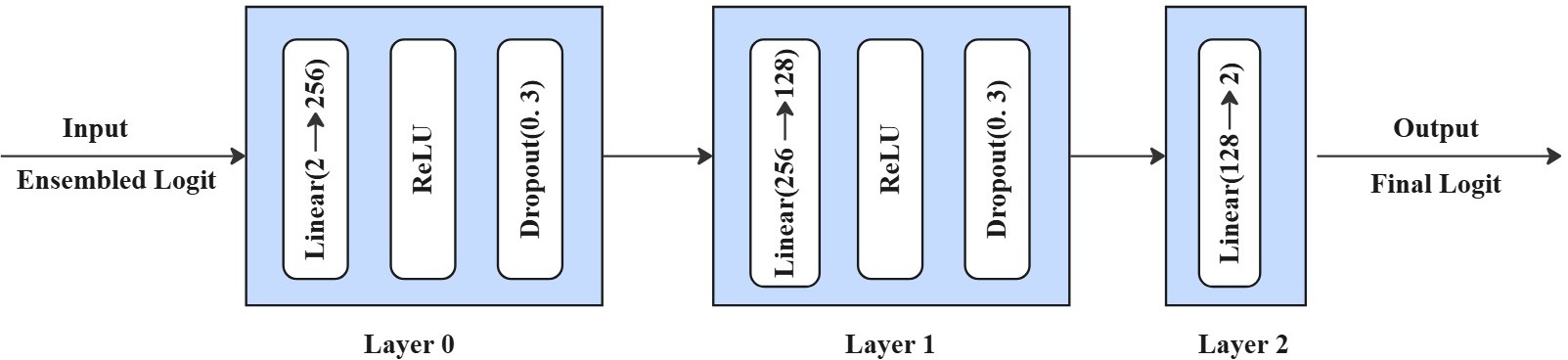}
    \caption{Architecture of the proposed Meta-MLP ensemble model used to combine teacher logits.}
    \label{fig:ensemble_architecture}
\end{figure*}

Figure~\ref{fig:ensemble_architecture} shows the proposed \textbf{Meta-Model Design}. The meta-model consists of three sequential layers. $Layer_0$ contains a fully connected dense layer of input dimension $2$, output dimension $256$, activation function ReLU, and dropout rate of $0.3$. $Layer_1$ contains another fully connected dense layer of input dimension $256$, output dimension $128$, activation function ReLU, and dropout rate of $0.3$. $Layer_2$ contains the final fully connected dense layer of input dimension $128$ and output dimension $2$.

\begin{figure*}[!t]
    \centering
    \includegraphics[width=\textwidth]{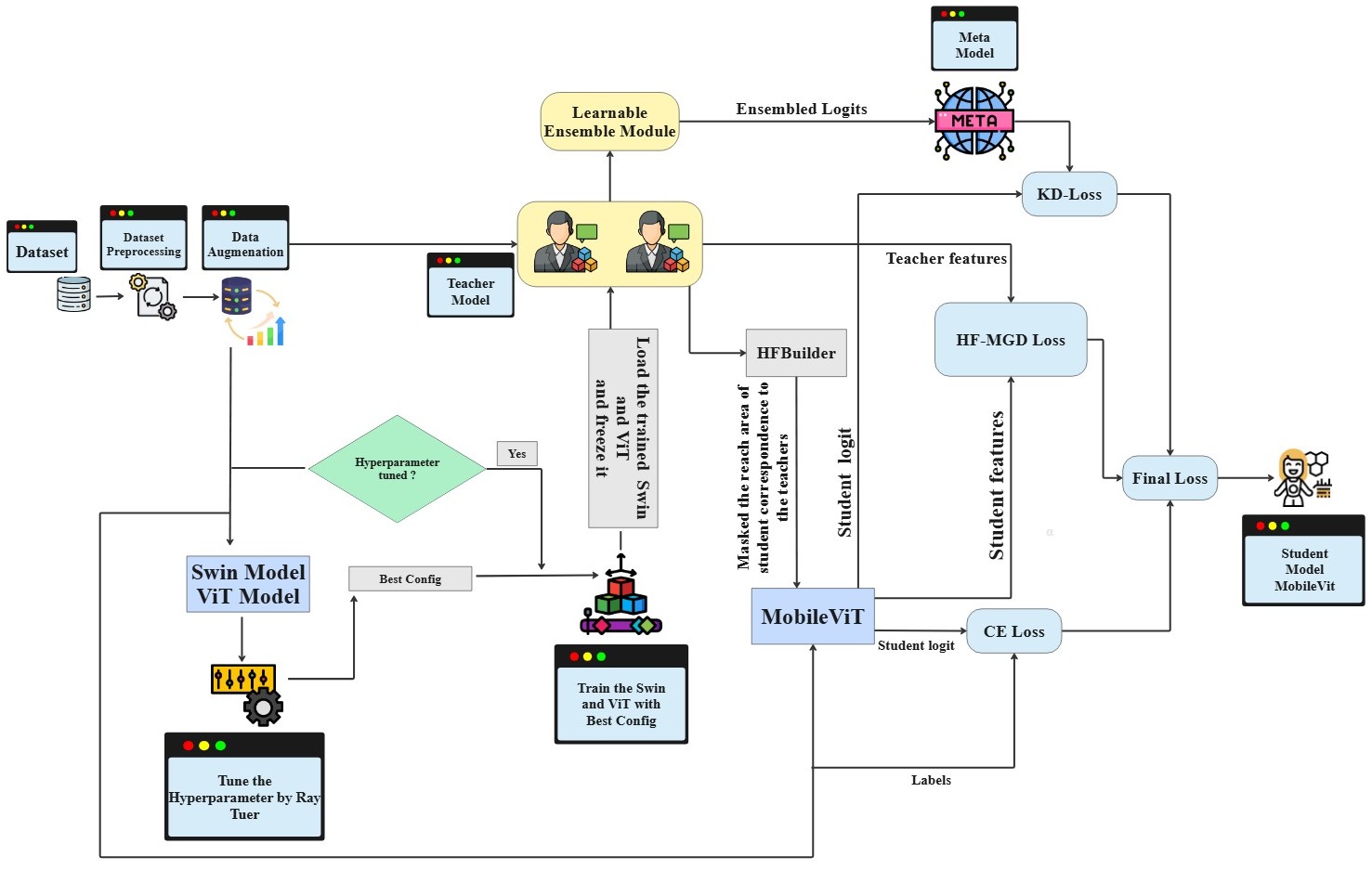}
    \caption{Proposed HumP-KD method, illustrating the interaction between teacher models, meta-model, and student model with integrated UAKD, HPKD, and MSKD modules.}
    \label{fig:overall-training-process}
\end{figure*}

Figure~\ref{fig:overall-training-process} shows the overall workflow, where Swin and ViT are the teacher models, the meta-model acts as the supervisor model, and MobileViT is the student model.

\begin{figure*}[!t]
    \centering
    \includegraphics[width=\textwidth]{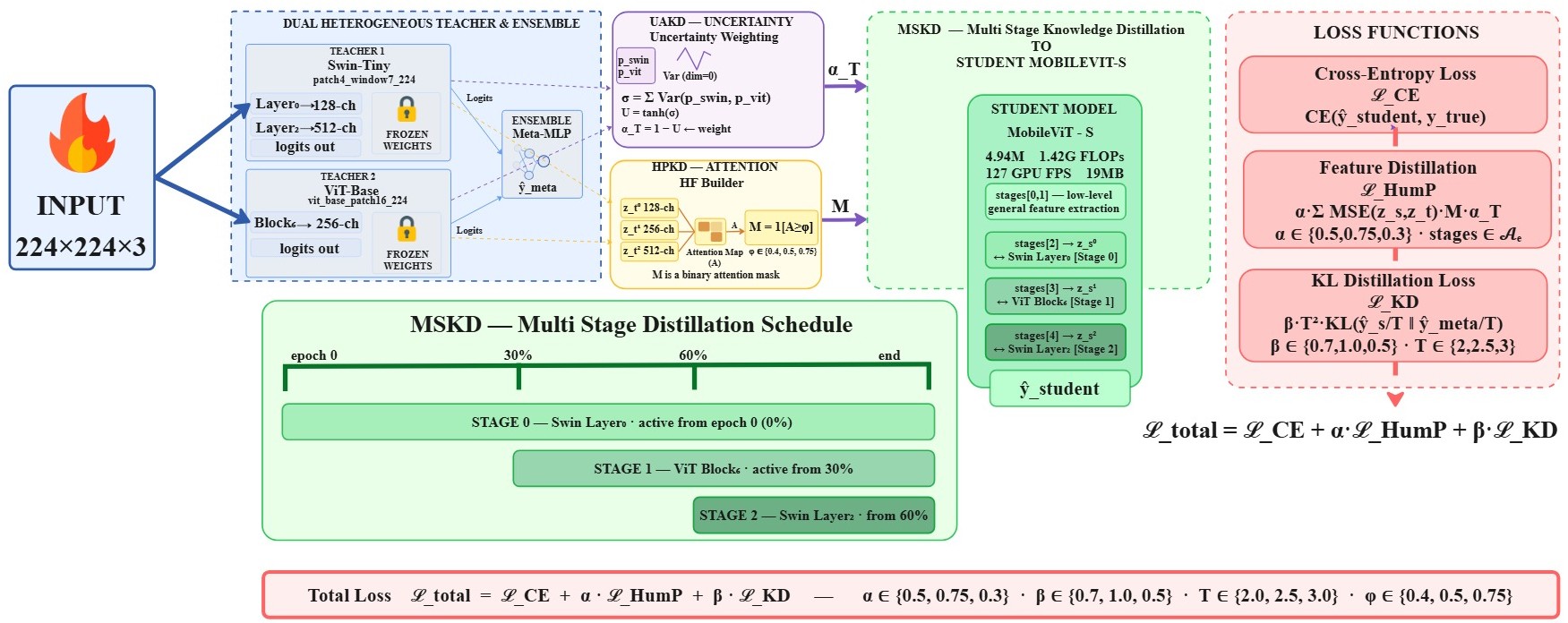}
    \caption{Proposed HumP-KD framework, illustrating dual-teacher knowledge distillation with UAKD, HPKD, and MSKD.}
    \label{fig:training-process-of-HumP-KD}
\end{figure*}

Figure~\ref{fig:training-process-of-HumP-KD} presents the full training process with heterogeneous teacher models, Swin-Tiny and ViT-Base, whose logits are passed through a frozen Meta-MLP ensemble to produce $\hat{y}_{meta}$. The Uncertainty-Aware Knowledge Distillation (UAKD) module computes a per-sample teacher confidence weight $\alpha_T$ based on the variance between teacher predictions. The Hierarchical Feature Builder (HFBuilder) fuses three teacher feature maps to generate a binary attention mask in (\ref{eq:binary_attention}), which selectively guides the Multi-Stage Knowledge Distillation (MSKD) across three progressive stages: Stage~0 from epoch~0, Stage~1 from 30\%, and Stage~2 from 60\%. The total training objective combines cross-entropy loss $\mathcal{L}_{CE}$, feature distillation loss $\mathcal{L}_{HumP}$, and KL distillation loss.

\subsection{Uncertainty-Aware Knowledge Distillation (UAKD)}
\label{uakd}

The predictive uncertainty of the teacher ensemble is quantified by computing the variance across the softmax outputs of the two teachers~\cite{10795207}. For a given input $\boldsymbol{x}$, the per-class probability outputs are:
\begin{equation}
\boldsymbol{\mathcal{P}}_{\text{swin}} = \text{softmax}(f_{\text{swin}}(\boldsymbol{x})), \quad
\boldsymbol{\mathcal{P}}_{\text{vit}} = \text{softmax}(f_{\text{vit}}(\boldsymbol{x}))
\end{equation}

The ensemble variance is computed as:
\begin{equation}
\boldsymbol{\sigma}^2 = \text{Var}\left([\boldsymbol{\mathcal{P}}_{\text{swin}},\ \boldsymbol{\mathcal{P}}_{\text{vit}}],\ \text{dim}=0\right)
\end{equation}

A per-sample scalar uncertainty is then derived as $\boldsymbol{\sigma} = \sum_c \boldsymbol{\sigma}_c^2$, and mapped to a confidence weight through:
\begin{equation}
\mathcal{U} = \tanh(\boldsymbol{\sigma}), \quad \alpha_T = 1 - \mathcal{U}
\end{equation}
where $\alpha_T$ represents teacher confidence, which is higher when teachers agree and lower when they disagree.

\subsection{Hierarchical Progressive Knowledge Distillation (HPKD)}

A Hierarchical Feature Builder $\boldsymbol{\mathcal{H}}$ takes the three teacher feature maps and produces a spatial attention map $\boldsymbol{A} = \mathcal{H}(\mathcal{Z}_t^0,\ \mathcal{Z}_t^1,\ \mathcal{Z}_t^2) \in \mathbb{R}^{B \times 1 \times H \times W}$. A binary attention mask is then defined as:
\begin{equation}
    \label{eq:binary_attention}
    \boldsymbol{M} = \mathbf{1}\left[\boldsymbol{A} \geq \phi\right]
\end{equation}
where $\phi$ is a threshold hyperparameter that controls how selective the distillation is in focusing on high-priority spatial regions.

\subsection{Multi-Stage Knowledge Distillation (MSKD)}
\label{mskd}

Stage activation follows a schedule based on training progress, where $p = e/E$, $e$ is the current epoch, and $E$ is the total number of epochs \cite{app12199453}. Stage $i$ becomes active when $p \geq \mathcal{T}_i$, where $\{\mathcal{T}_0, \mathcal{T}_1, \mathcal{T}_2\} = \{0.0, 0.3, 0.6\}$. The active stage set at epoch $e$ is defined as:
\begin{equation}
    \mathcal{A}_e = \{i \mid p \geq \mathcal{T}_i\}, \quad i \in \{0, 1, 2\}
\end{equation}
For each stage $i \in \mathcal{A}_e$, we project the student features through an adapter $\psi_i$ and a projector $\rho_i$, mask them by the interpolated attention map $\boldsymbol{M}_i$, and align them with normalized teacher features. The per-stage distillation MSE loss is averaged over the spatial and channel dimensions, weighted by the teacher confidence $\alpha_T$, and then averaged across the batch.

\subsection{Distillation Setup}

\textbf{Learnable Ensemble.}
This component has two scalar tensors, $w_{\text{swin}}$ and $w_{\text{vit}}$, with a default value of $0.5$. After dimension reduction, softmax is applied to both Swin and ViT outputs, assigning the results to $\boldsymbol{p}_{\text{swin}}$ and $\boldsymbol{p}_{\text{vit}}$. The ensemble weight is defined as:
\begin{equation}
    \hat{\boldsymbol{y}}_{\text{ens}} =
    \frac{w_{\text{swin}}}{w_{\text{swin}} + w_{\text{vit}}}
    \boldsymbol{p}_{\text{swin}} +
    \frac{w_{\text{vit}}}{w_{\text{swin}} + w_{\text{vit}}}
    \boldsymbol{p}_{\text{vit}}
\end{equation}

The standard knowledge distillation loss is defined as:
\begin{align}
    \boldsymbol{s} &= \hat{\boldsymbol{y}}_s / T \\
    \boldsymbol{t} &= \hat{\boldsymbol{y}}_{\text{meta}} / T
\end{align}
\begin{align}
    \boldsymbol{s}_{\text{plog}} &= \log\text{-softmax}(\boldsymbol{s}) \\
    \boldsymbol{t}_{p} &= \text{softmax}(\boldsymbol{t})
\end{align}
\begin{equation}
    \mathcal{L}_{\text{KD}} = T^2 \cdot \mathcal{L}_{\text{KL}}\left(\boldsymbol{s}_{\text{plog}},\ \boldsymbol{t}_{p}\right)
\end{equation}

The entropy from the teacher logits for uncertainty awareness is defined as:
\begin{equation}
    \boldsymbol{p} = \text{softmax}(\hat{\boldsymbol{y}}_t)
\end{equation}
\begin{equation}
    \mathcal{H} = -\frac{1}{N}\sum_{i=1}^{N}\sum_{j=1}^{C} p_{i,j} \log\left(p_{i,j}\right)
\end{equation}

We load the student model MobileViT-small with average global pooling and extract the three hierarchical stages $\{\mathcal{Z}_s^0,\ \mathcal{Z}_s^1,\ \mathcal{Z}_s^2\}$ via forward hooks registered at stages $\{2, 3, 4\}$. We also load the pre-trained teacher models $f_{\text{swin}}$, $f_{\text{vit}}$, and the Meta Model $f_{\text{meta}}$, and extract the Swin layers $\{0, 1, 2\}$ and ViT blocks $\{3, 6, 9\}$. We define the Hierarchical Feature Builder $\boldsymbol{HFBuilder}$ that generates a hierarchical attention map from the teachers.

\textbf{HFBuilder.}
It acts as a hierarchical feature aggregator that bridges the structural gap between the teachers' high-resolution and low-resolution layers and blocks. It contains a module $reduce$ with $1\times1$ convolutions that transform feature depths $(128, 256, 512)$ into a consistent $64$-channel format, two scalar parameters $\gamma_2$ and $\gamma_3$, and a final convolution $finalconv$ with output channel $1$.

\begin{align}
\texttt{reduce} &: \mathbb{R}^{B \times C_{\text{in}} \times H \times W} \rightarrow \mathbb{R}^{B \times 64 \times H \times W}, \quad C_{\text{in}} \in \{128, 256, 512\} \\
\texttt{finalconv} &: \mathbb{R}^{B \times 64 \times H \times W} \rightarrow \mathbb{R}^{B \times 1 \times H \times W}
\end{align}

The function \texttt{to\_hw} reconstructs the 2D grid from the 1D sequence. For $\boldsymbol{f}_1 = \mathcal{Z}_t^0$, $\boldsymbol{f}_2 = \mathcal{Z}_t^1$, and $\boldsymbol{f}_3 = \mathcal{Z}_t^2$, the features are passed through $reduce$ and fused as:
\begin{equation}
\boldsymbol{f}_{\text{fuse}} = \boldsymbol{r}_1 + \gamma_2 \boldsymbol{r}_2 + \gamma_3 \boldsymbol{r}_3
\end{equation}

The salience map is defined as:
\begin{equation}
\boldsymbol{A} = \text{sigmoid}\left( finalconv\left(\text{GELU}\left(\boldsymbol{f}_{\text{fuse}}\right)\right)\right)
\end{equation}

We define $\mathcal{C}_t = \{128, 256, 512\}$ and $\mathcal{C}_s$ as the student feature dimensions at stages $\{2, 3, 4\}$. The adapter $\psi$ consists of three $1\times1$ convolutional networks:
\begin{equation}
\psi_i : \mathbb{R}^{B \times \mathcal{C}_s^i \times H \times W} \rightarrow \mathbb{R}^{B \times \mathcal{C}_t^i \times H \times W}, \quad i \in \{0, 1, 2\}
\end{equation}

The student projector $\rho_s$ is defined as:
\begin{equation}
\rho_s^i : \mathbb{R}^{B \times \mathcal{C}_t^i} \rightarrow \mathbb{R}^{B \times \mathcal{C}_t^i}, \quad i \in \{0, 1, 2\}
\end{equation}

We define the hyperparameters $\alpha = 0.5,\ \beta = 0.7,\ \phi = 0.5,\ T = 2$, and the $HumP-KD$ loss function shown in Algorithm~\ref{alg:HumP-KD}. The teacher projectors $\rho_t$ map each teacher feature to a consistent channel dimension:
\begin{align}
\boldsymbol{\mathcal{\rho}}_t^0 &: \mathbb{R}^{B \times C_{\text{swin}}^0 \times H \times W} \rightarrow \mathbb{R}^{B \times 128 \times H \times W} \\
\boldsymbol{\mathcal{\rho}}_t^1 &: \mathbb{R}^{B \times N \times C_{\text{vit}}^6} \rightarrow \mathbb{R}^{B \times N \times 256} \\
\boldsymbol{\mathcal{\rho}}_t^2 &: \mathbb{R}^{B \times C_{\text{swin}}^2 \times H \times W} \rightarrow \mathbb{R}^{B \times 512 \times H \times W}
\end{align}

Algorithm~\ref{alg:HumP-KD} summarizes the complete computation of the HumP-KD distillation loss, integrating \textbf{HPKD}, \textbf{UAKD}, and \textbf{MSKD} into a single unified procedure.

\begin{algorithm}[H]
\caption{HumP-KD: Hierarchical Uncertainty-aware Multi-stage Progressive Distillation Loss}
\label{alg:HumP-KD}
\begin{algorithmic}[1]
\Require Student features $\{s_0, s_1, s_2\}$, teacher features $\{t_0, t_1, t_2\}$,
teacher logits $\hat{y}_{\text{swin}},\ \hat{y}_{\text{vit}}$, current epoch $e$, total epochs $E$,
adapters $\{\psi_i\}$, projectors $\{\rho_s^i\}$, HF Builder $hf_{builder}$,
attention threshold $\phi$, stage thresholds $\{\mathcal{T}_0, \mathcal{T}_1, \mathcal{T}_2\} = \{0.0, 0.3, 0.6\}$
\Ensure Distillation loss $\mathcal{L}_{\text{HumP}}$

\State $p \leftarrow e / E$
\State $\mathcal{A}_e \leftarrow \{i \mid p \geq \mathcal{T}_i,\ i \in \{0, 1, 2\}\}$
\If{$\mathcal{A}_e = \emptyset$}
    \State \Return $0$
\EndIf

\State $A \leftarrow hf_{builder}(t_0,\ t_1,\ t_2)$
\State $M \leftarrow \mathbf{1}[A \geq \phi]$

\State $\mathcal{P}_{\text{swin}} \leftarrow \text{softmax}(\hat{y}_{\text{swin}})$
\State $\mathcal{P}_{\text{vit}} \leftarrow \text{softmax}(\hat{y}_{\text{vit}})$
\State $\sigma^2 \leftarrow \text{Var}([\mathcal{P}_{\text{swin}},\ \mathcal{P}_{\text{vit}}],\ \text{dim}=0)$
\State $\sigma \leftarrow \sum_c \sigma_c^2$
\State $\mathcal{U} \leftarrow \tanh(\sigma)$
\State $\alpha_T \leftarrow 1 - \mathcal{U}$

\State $\mathcal{L} \leftarrow 0$
\For{each active stage $i \in \mathcal{A}_e$}
    \State $s \leftarrow \psi_i(s_i)$
    \State $M_i \leftarrow \text{Interpolate}(M,\ (H_i, W_i))$
    \State $s \leftarrow \text{Flatten}(s)^\top$
    \State $s \leftarrow s \odot \text{Flatten}(M_i)^\top$
    \State $s \leftarrow \rho_s^i(s)$
    \State $N \leftarrow \min(|s|_{\text{seq}},\ |t_i|_{\text{seq}})$
    \State $\tilde{s} \leftarrow \text{Normalize}(s_{:,:N},\ \text{dim}=-1)$
    \State $\tilde{t}_i \leftarrow \text{Normalize}(t_{i,:,:N},\ \text{dim}=-1)$
    \State $\mathcal{L}_i \leftarrow \frac{1}{NC}\sum_{n,c}(\tilde{s}_{n,c} - \tilde{t}_{i,n,c})^2$
    \State $\mathcal{L}_i \leftarrow \frac{1}{B}\sum_b \alpha_T^{(b)} \cdot \mathcal{L}_i^{(b)}$
    \State $\mathcal{L} \leftarrow \mathcal{L} + \mathcal{L}_i$
\EndFor

\State \Return $\mathcal{L}_{\text{HumP}} \leftarrow \mathcal{L} / |\mathcal{A}_e|$
\end{algorithmic}
\end{algorithm}

We pass the parameters of the student ($f_s$), adapters ($\psi$), both student and teacher projectors ($\rho_s,\rho_t$), ensemble weights, and attention generator to the Adam optimizer.

\textbf{Training.}
For each mini-batch $(\boldsymbol{x}, y) \sim \mathcal{D}_{\text{train}}$, the teacher models perform a frozen forward pass:
\begin{equation}
\hat{\boldsymbol{y}}_{\text{swin}} = f_{\text{swin}}(\boldsymbol{x}), \quad 
\hat{\boldsymbol{y}}_{\text{vit}} = f_{\text{vit}}(\boldsymbol{x})
\end{equation}
\begin{equation}
\boldsymbol{t}_i = \boldsymbol{\mathcal{\rho}}_t^i\left(\mathcal{Z}_t^i\right), \quad i \in \{0, 1, 2\}
\end{equation}

The meta model produces a refined ensemble prediction:
\begin{equation}
\hat{\boldsymbol{y}}_{\text{ens}} = \texttt{ensemble\_weights}\left(\hat{\boldsymbol{y}}_{\text{swin}},\ \hat{\boldsymbol{y}}_{\text{vit}}\right)
\end{equation}
\begin{equation}
\hat{\boldsymbol{y}}_{\text{meta}} = f_{\text{meta}}\left(\hat{\boldsymbol{y}}_{\text{ens}}\right)
\end{equation}

The student produces logits $\hat{\boldsymbol{y}}_s = f_s(\boldsymbol{x})$ and features $\{\mathcal{Z}_s^0, \mathcal{Z}_s^1, \mathcal{Z}_s^2\}$. The training loss is:
\begin{equation}
\mathcal{L}_{\text{train}} = \mathcal{L}_{\text{CE}}(\hat{\boldsymbol{y}}_s, y) + \alpha \cdot \mathcal{L}_{\text{HumP}} + \beta \cdot \mathcal{L}_{\text{KD}}(\hat{\boldsymbol{y}}_s,\ \hat{\boldsymbol{y}}_{\text{meta}},\ T)
\end{equation}

The parameters $\Theta$ are updated as:
\begin{equation}
\Theta \leftarrow \Theta - \eta \cdot \nabla_{\Theta}\, \mathcal{L}_{\text{train}}, \quad \eta = 1\times10^{-4}
\end{equation}

\textbf{Validation.}
For each mini-batch $(\boldsymbol{x}, y) \sim \mathcal{D}_{\text{val}}$, the student logits $\hat{\boldsymbol{y}}_s = f_s(\boldsymbol{x})$ are evaluated. The validation loss excludes $\mathcal{L}_{\text{HumP}}$:
\begin{equation}
\mathcal{L}_{\text{val}} = \mathcal{L}_{\text{CE}}(\hat{\boldsymbol{y}}_s, y) + \beta \cdot \mathcal{L}_{\text{KD}}(\hat{\boldsymbol{y}}_s,\ \hat{\boldsymbol{y}}_{\text{meta}},\ T)
\end{equation}

The best student checkpoint is saved when validation accuracy improves:
\begin{equation}
\hat{\theta}_{f_s} = \underset{\theta_{f_s}}{\arg\max}\ \text{Acc}_{\text{val}}(e), \quad e \in \{1, \ldots, E\}
\end{equation}

The overall loss is:
\begin{equation}
\mathcal{L}_{\text{total}} = \mathcal{L}_{\text{CE}}(\hat{\boldsymbol{y}}_s, y) + \alpha \cdot \mathcal{L}_{\text{HumP}} + \beta \cdot \mathcal{L}_{\text{KD}}(f_s(\boldsymbol{x}),\ \hat{\boldsymbol{y}}_{\text{meta}},\ T)
\end{equation}
\begin{equation}
\mathcal{L}_{\text{KD}} = T^2 \cdot D_{\text{KL}}\left(\log\text{-softmax}\!\left(\frac{\hat{\boldsymbol{y}}_s}{T}\right) \,\Big\|\, \text{softmax}\!\left(\frac{\hat{\boldsymbol{y}}_{\text{meta}}}{T}\right)\right)
\end{equation}

The objective is to minimize $\mathcal{L}_{\text{total}}$ over all parameters of $f_s$, $\{\psi_i\}$, $\{\rho^i\}$, and $\boldsymbol{hf_{builder}}$, such that the student achieves competitive fire classification accuracy while maintaining a compact footprint of $4.94$M parameters, $1.42$G FLOPs, and real-time inference speed exceeding $127$ GPU FPS.

\section{Results and Discussion}
In this section, we present the detailed results of applied deep learning models for fire detection. We have analyzed the performance of various models using different metrics and optimization techniques. All our experiments were conducted on a workstation equipped with NVIDIA GeForce RTX 3090 GPU with 24GB VRAM.


\begin{table*}[t]
\centering
\caption{Optimized hyperparameter settings for various models}
\label{tab:hyperparameters_table}
\renewcommand{\arraystretch}{1.2}
\begin{tabular}{lccccccc}
\toprule

Model & Learning Rate & Dropout & Optimizer & Weight Decay & Regularization & Reg. Value & Epochs \\

\midrule

MobileNetV2 & 1e-4 & 0.3 & Adam & 0.0 & L1 & 1e-4 & 10 \\
EfficientNetB0 & 1e-2 & 0.6 & Adam & 0.0 & L1 & 1e-4 & 10 \\
EfficientNetB3 & 1e-3 & 0.5 & Adam & 0.0 & L1 & 1e-4 & 10 \\
Xception & 1e-3 & 0.4 & Adam & 0.0 & L1 & 1e-5 & 10 \\
Xception + ResNet50 & 5e-4 & 0.5 & AdamW & 1e-4 & L1 & 1e-4 & 10 \\
ViT-Base & 1e-2 & 0.4 & RMSprop & 1e-4 & L1 & 1e-5 & 10 \\
DeiT-III & 5.23e-4 & 0.3 & Adam & 0.0 & L1 & 1e-6 & 10 \\
Swin-Tiny & 5.23e-4 & 0.4 & SGD & 1e-6 & None & -- & 10 \\
Swin-Small & 4.01e-4 & 0.3 & Adam & 0.0 & None & -- & 10 \\
MobileViT-S & 6.02e-4 & 0.2 & AdamW & 0.0 & None & -- & 10 \\
\bottomrule
\end{tabular}
\end{table*}

Table~\ref{tab:hyperparameters_table} demonstrates the optimized hyperparameter settings obtained using automatic \textbf{Ray Tuner} for all evaluated models. The \textbf{ASHA Scheduler} was employed to enable efficient early stopping by pruning underperforming trials during the tuning process.

\begin{figure}[h]
    \centering
    \includegraphics[width=1.0\linewidth]{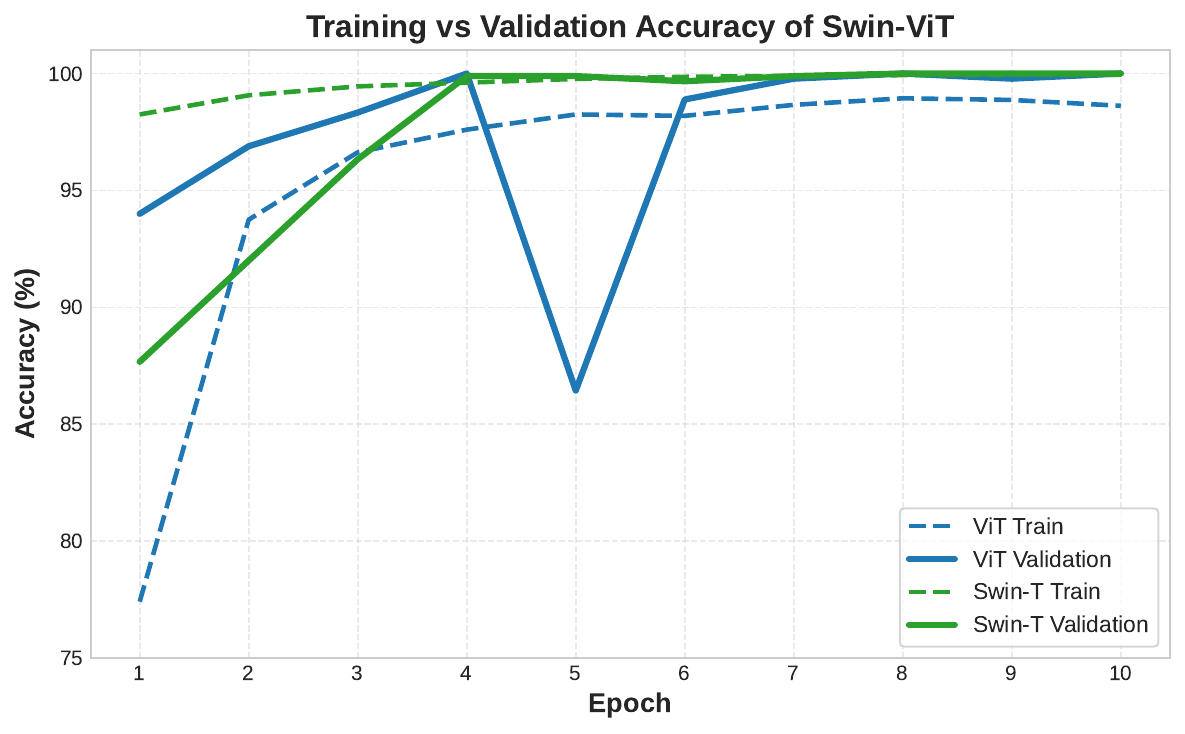}    
    \caption{Training and validation accuracy curves of ViT and Swin-T.}
    \label{fig:accuracy_comparison_swin_vit}    
\end{figure}

Figure~\ref{fig:accuracy_comparison_swin_vit} shows that Swin-T achieves a higher and more stable validation accuracy, reaching approximately 96--98\%, compared to ViT, which fluctuates around 90--94\%.

\begin{figure}[htb]
    \centering
    \includegraphics[width=1.0\linewidth]{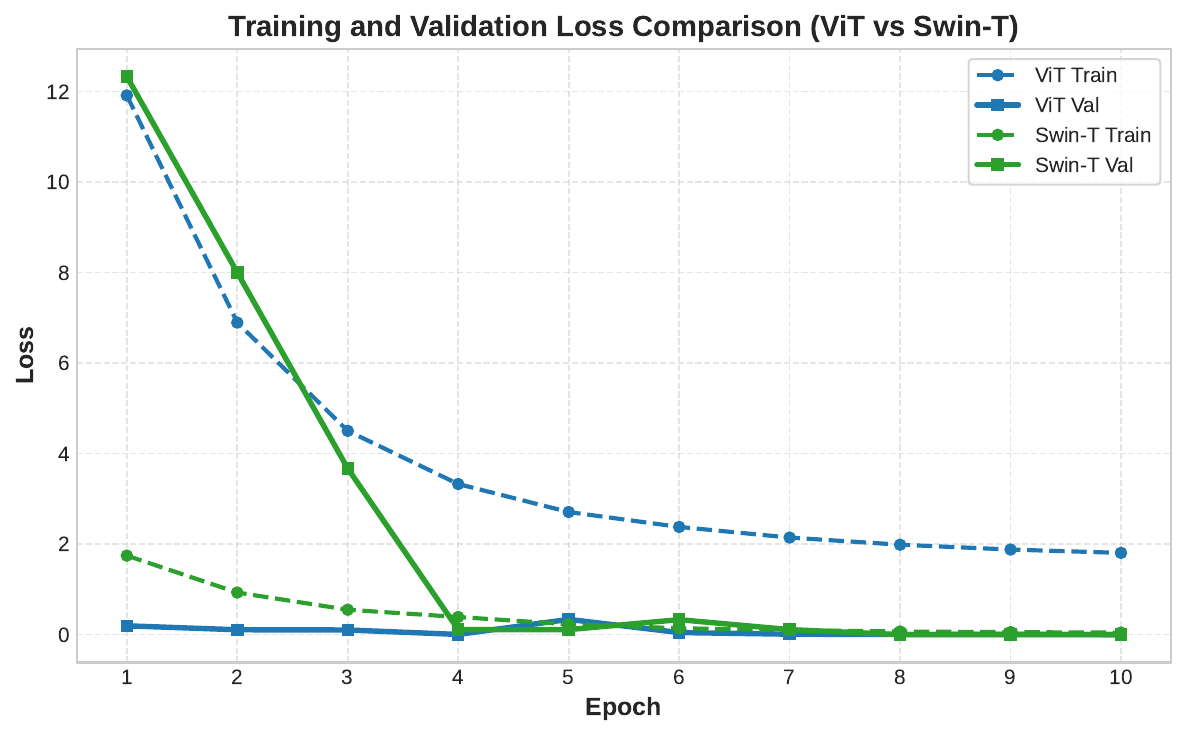}    
    \caption{Training and validation loss curves of ViT and Swin-T.}
    \label{fig:loss_comparison_swin_vit}    
\end{figure}

According to Figure~\ref{fig:loss_comparison_swin_vit}, the Swin-T model converges faster, with its validation loss stabilizing by around epoch 3–4, whereas ViT continues to fluctuate until later epochs. Swin-T maintains a lower and more consistent validation loss (approximately below 0.1) compared to ViT.

\begin{table*}[t]
\centering
\caption{Performance of various models under different settings on Dataset-II (Default vs. Optimized hyperparameters, with and without augmentation)}
\label{tab:merged_performance}
\setlength{\tabcolsep}{4pt}
\renewcommand{\arraystretch}{1.2}

\begin{tabular}{lcccccccc}
\toprule

& \multicolumn{4}{c}{\textbf{Default Hyperparameters}} & \multicolumn{4}{c}{\textbf{Optimized Hyperparameters}} \\

\cmidrule(lr){2-5} \cmidrule(lr){6-9}

\textbf{Model} 
& \textbf{No Aug} & \textbf{Aug} & \textbf{No Aug} & \textbf{Aug} 
& \textbf{No Aug} & \textbf{Aug} & \textbf{No Aug} & \textbf{Aug} \\

& \multicolumn{2}{c}{\textbf{Accuracy (\%)}} & \multicolumn{2}{c}{\textbf{F1 Score (\%)}} 
& \multicolumn{2}{c}{\textbf{Accuracy (\%)}} & \multicolumn{2}{c}{\textbf{F1 Score (\%)}} \\

\midrule

MobileNetV2      
& 89.98 & 92.88 & 90.57 & 93.21 
& 100.00 & 98.11 & 100.00 & 98.14 \\

EfficientNetB0   
& 99.25 & 94.42 & 99.21 & 94.24 
& 64.78 & 99.56 & 67.41 & 99.55 \\

EfficientNetB3   
& 99.34 & 97.54 & 99.33 & 97.50 
& 94.67 & 100.00 & 94.65 & 100.00 \\

Xception         
& 100.00 & 99.67 & 100.00 & 99.67 
& 100.00 & 99.89 & 100.00 & 99.89 \\

Xception+ResNet50 
& 100.00 & 99.56 & 100.00 & 99.56 
& 100.00 & 98.56 & 100.00 & 98.57 \\

ViT              
& 98.78 & 99.33 & 98.79 & 99.34 
& 99.33 & 100.00 & 99.34 & 100.00 \\

DeiT-III         
& 97.99 & 100.00 & 97.99 & 100.00 
& 98.41 & 99.22 & 98.41 & 99.22 \\

Swin-Tiny        
& 99.56 & 99.56 & 99.56 & 99.56 
& 100.00 & 100.00 & 100.00 & 100.00 \\

Swin-Small       
& 100.00 & 98.33 & 100.00 & 98.33 
& 99.55 & 100.00 & 99.55 & 100.00 \\

MobileViT-S      
& 100.00 & 99.56 & 100.00 & 99.56 
& 99.78 & 100.00 & 99.78 & 100.00 \\

\bottomrule
\end{tabular}
\end{table*}

Table~\ref{tab:merged_performance} compares model performance under default and optimized hyperparameters, with and without augmentation. Under default settings, several models, such as Xception and MobileViT-S, achieve up to 100.00\% accuracy and F1 score without augmentation, while MobileNetV2 performs lower at 89.98\% accuracy and 90.57\% F1 score. With optimized hyperparameters and augmentation, most models show improved and consistent performance, with EfficientNetB3, ViT, Swin-Tiny, and MobileViT-S reaching 100.00\% accuracy and F1 score. Certain models, such as EfficientNetB0, exhibit instability under optimized settings without augmentation, dropping to 64.78\% accuracy.

\subsection{Ablation Study}

\begin{table*}[t]
\centering
\caption{Ablation study of HumP-KD showing the contribution of different components and teacher layer importance}
\label{tab:ablation_table}
\renewcommand{\arraystretch}{1.2}

\begin{tabular}{lcccccccccc}
\toprule
& \multicolumn{2}{c}{HPKD} & \multicolumn{3}{c}{Components} & \multicolumn{3}{c}{Teacher Layer Importance} & \multicolumn{2}{c}{Performance} \\
\cmidrule(lr){2-3}
\cmidrule(lr){4-6}
\cmidrule(lr){7-9}
\cmidrule(lr){10-11}
Model & Attention & Meta Model & Augmentation & UAKD & MSKD & L0 & B6 & L2 & Acc (\%) & F1 Score \\
\midrule
HumP-KD & $\checkmark$ & $\checkmark$ & $\checkmark$ & $\times$ & $\times$ & $\checkmark$ & $\checkmark$ & $\checkmark$ & 98.63 & 0.9863 \\
HumP-KD & $\times$ & $\checkmark$ & $\checkmark$ & $\checkmark$ & $\checkmark$ & -- & -- & -- & 97.33 & 0.9733 \\
HumP-KD & $\checkmark$ & $\checkmark$ & $\times$ & $\times$ & $\times$ & $\checkmark$ & $\checkmark$ & $\checkmark$ & 98.64 & 0.9864 \\
HumP-KD & $\checkmark$ & $\checkmark$ & $\checkmark$ & $\times$ & $\checkmark$ & $\times$ & $\times$ & $\times$ & 96.31 & 0.9631 \\
HumP-KD & $\checkmark$ & $\times$ & $\checkmark$ & $\checkmark$ & $\times$ & $\checkmark$ & $\checkmark$ & $\checkmark$ & 94.33 & 0.9433 \\
HumP-KD & $\checkmark$ & $\checkmark$ & $\checkmark$ & $\checkmark$ & $\checkmark$ & $\checkmark$ & $\checkmark$ & $\checkmark$ & 98.48 & 0.9848 \\
\bottomrule
\end{tabular}
\end{table*}

Table~\ref{tab:ablation_table} shows the importance of each component in improving the robustness of the student model. When the attention component is removed, the teacher layers are not required (indicated by "--"), and the performance drops to 97.33\% accuracy and 0.9733 F1 score. To demonstrate the importance of Swin-T's $layer_0$, $layer_2$, and ViT's $block_6$, we replaced them with $layer_1$, $layer_3$, and $block_9$, resulting in a performance drop from 98.48\% to 96.31\% accuracy. Removing additional components generally leads to performance degradation, as seen when excluding the meta model, which reduces performance to 94.33\% accuracy and 0.9433 F1 score. For augmentation, the model performs slightly better without it, achieving 98.64\% accuracy compared to 98.48\%. As the teacher models perform strongly on the FlameVision dataset, the ablation study is conducted on Dataset-II.

\begin{table*}[t]
\centering
\caption{Computational cost and inference speed comparison of CNN and transformer models}
\label{tab:computational_cost}
\begin{tabularx}{\textwidth}{lcccccc}
\toprule
\textbf{Model} & \textbf{Total Params (M)} & \textbf{Trainable Params (M)} & \textbf{Model Size (MB)} & \textbf{FLOPs (G)} & \textbf{GPU FPS} & \textbf{CPU FPS} \\
\midrule
MobileNetV2        & 2.39  & 2.15  & 9.35   & 0.33  & 236.79 & 88.16 \\
EfficientNetB0     & 4.01  & 4.01  & 15.59  & 0.41  & 133.58 & 31.55 \\
EfficientNetB3     & 10.70 & 10.69 & 41.35  & 1.02  & 80.30  & 31.55 \\
Xception           & 21.07 & 21.07 & 80.67  & 4.60  & 21.07  & 24.16 \\
Xception + ResNet50 & 44.32 & 0.01  & 169.69 & 8.73  & 115.01 & 19.21 \\
ViT-Base           & 86.19 & 86.19 & 328.87 & 16.85 & 214.23 & 15.59 \\
DeiT-III           & 86.21 & 86.21 & 328.94 & 16.85 & 196.05 & 15.68 \\
Swin-Tiny          & 27.91 & 27.91 & 106.56 & 4.37  & 115.82 & 28.19 \\
Swin-Small         & 49.23 & 49.23 & 187.93 & 8.54  & 56.82  & 15.56 \\
\textbf{MobileViT-S (HumP-KD)} & \textbf{4.94} & \textbf{4.94} & \textbf{19.01} & \textbf{1.42} & \textbf{134.57} & \textbf{37.72} \\
\bottomrule
\end{tabularx}
\end{table*}

Table~\ref{tab:computational_cost} shows that the proposed MobileViT-S (HumP-KD) achieves strong efficiency with only 4.94M parameters and 1.42 GFLOPs, while maintaining a compact model size of 19.01 MB. It also delivers 134.57 GPU FPS and 37.72 CPU FPS, significantly outperforming larger models such as ViT (86.19M parameters, 16.85 GFLOPs) in real-time deployment suitability.

\begin{table*}[t]
\centering
\caption{Comparison of HumP-KD with existing knowledge distillation methods under clean and corrupted (Gaussian noise + motion blur) conditions}
\label{tab:kd_comparison_merged}
\setlength{\tabcolsep}{5pt}
\renewcommand{\arraystretch}{1.2}

\begin{tabular}{llcccccccc}
\toprule

& & \multicolumn{4}{c}{\textbf{Clean Data}} & \multicolumn{4}{c}{\textbf{Corrupted Data}} \\

\cmidrule(lr){3-6} \cmidrule(lr){7-10}

\textbf{Model} & \textbf{Dataset} 
& \textbf{Acc} & \textbf{Prec} & \textbf{Rec} & \textbf{F1} 
& \textbf{Acc} & \textbf{Prec} & \textbf{Rec} & \textbf{F1} \\

\midrule

\multirow{2}{*}{Vanilla-KD}
& FlameVision & 100.00 & 100.00 & 100.00 & 100.00 & 95.93 & 94.73 & 95.93 & 95.93 \\
& Dataset-II  & 81.40  & 83.41  & 81.40  & 81.15  & 77.73 & 79.21 & 77.75 & 78.22 \\

\midrule

\multirow{2}{*}{DIST}
& FlameVision & 78.67 & 83.26 & 78.67 & 70.11 & 100.00 & 100.00 & 100.00 & 100.00 \\
& Dataset-II  & 98.46 & 98.46 & 98.46 & 98.46 & 93.28 & 93.72 & 93.28 & 93.26 \\

\midrule

\multirow{2}{*}{ReviewKD}
& FlameVision & 100.00 & 100.00 & 100.00 & 100.00 & 100.00 & 100.00 & 100.00 & 100.00 \\
& Dataset-II  & 98.90 & 98.90 & 98.90 & 98.90 & 96.31 & 96.44 & 96.31 & 96.31 \\

\midrule

\multirow{2}{*}{CRD}
& FlameVision & 100.00 & 100.00 & 100.00 & 100.00 & 93.78 & 94.24 & 93.78 & 93.39 \\
& Dataset-II  & 98.67 & 98.67 & 98.67 & 98.67 & 96.98 & 97.03 & 96.98 & 96.98 \\

\midrule

\multirow{2}{*}{\textbf{HumP-KD}}
& FlameVision & 94.78 & 95.18 & 94.78 & 94.88 & 97.22 & 97.32 & 97.22 & 97.15 \\
& Dataset-II  & 98.48 & 98.50 & 98.48 & 98.48 & 97.62 & 97.64 & 97.62 & 97.62 \\

\bottomrule
\end{tabular}
\end{table*}

Table~\ref{tab:kd_comparison_merged} shows that while several methods achieve perfect performance on clean FlameVision data (100.00\%), their performance drops under corrupted conditions, such as CRD decreasing to 93.78\% accuracy. The proposed HumP-KD model demonstrates stronger robustness, improving from 94.78\% to 97.22\% accuracy on FlameVision and achieving 97.62\% accuracy on Dataset-II under noise and blur. This highlights the key contribution of HumP-KD in maintaining stable performance under degraded visual conditions rather than only optimizing peak accuracy on clean data.

\begin{figure}[htb]
    \centering
    \includegraphics[width=1.0\linewidth]{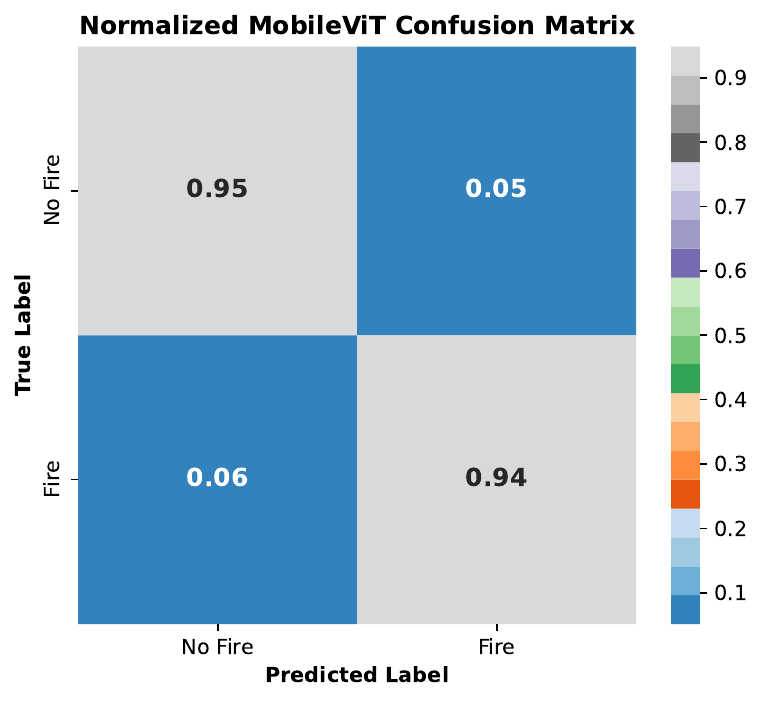}    
    \caption{Normalized confusion matrix of the HumP-KD student (MobileViT-S) on the FlameVision dataset.}
    \label{fig:student-mobilevit_confusion_matrix}    
\end{figure}

Figure~\ref{fig:student-mobilevit_confusion_matrix} shows that the model achieves high classification accuracy, correctly identifying 95\% of no-fire and 94\% of fire samples. Misclassification rates remain low, with only 5\% of no-fire and 6\% of fire instances incorrectly predicted.

\begin{table*}[t]
\centering
\caption{Cross-dataset generalization performance of HumP-KD between FlameVision and Dataset-II}
\label{tab:cross_dataset_testing}
\setlength{\tabcolsep}{6pt}
\renewcommand{\arraystretch}{1.2}
\begin{tabular}{lcccc}
\toprule
\multirow{3}{*}{\textbf{Metrics}} & \multicolumn{2}{c}{\textbf{HumP-KD: Trained on FlameVision and}} & \multicolumn{2}{c}{\textbf{HumP-KD: Trained on Dataset-II and}} \\
 & \multicolumn{2}{c}{\textbf{Test on Dataset-II}} & \multicolumn{2}{c}{\textbf{Test on FlameVision}} \\
\cmidrule(lr){2-3} \cmidrule(lr){4-5}
 & \textbf{Without Fine-tune} & \textbf{With Fine-tune} & \textbf{Without Fine-tune} & \textbf{With Fine-tune} \\
 & \textbf{on Dataset-II} & \textbf{on Dataset-II} & \textbf{on FlameVision} & \textbf{on FlameVision} \\
\midrule
Accuracy (\%)  & 88.09 & 96.86 & 86.22 & 98.00 \\
Precision (\%) & 88.40 & 96.94 & 91.50 & 98.14 \\
Recall (\%)    & 88.09 & 96.86 & 86.22 & 98.00 \\
F1 (\%)  & 88.06 & 96.86 & 87.18 & 98.03 \\
\bottomrule
\end{tabular} 
\end{table*}

Table~\ref{tab:cross_dataset_testing} shows that without fine-tuning, the model achieves 88.09\% accuracy when trained on FlameVision and tested on Dataset-II, and 86.22\% in the reverse setting. With fine-tuning, performance improves significantly to 96.86\% and 98.00\% accuracy, demonstrating strong cross-dataset generalization capability of HumP-KD.

\begin{table*}[t]
\centering
\caption{Performance of HumP-KD under different hyperparameter settings on FlameVision dataset}
\label{tab:HumP-KD_param_settings}
\setlength{\tabcolsep}{6pt}
\renewcommand{\arraystretch}{1.3}
\begin{tabular}{lcccc}
\toprule
\multirow{2}{*}{\textbf{Hyperparameter Settings}} & \multicolumn{4}{c}{\textbf{HumP-KD model trained and tested on FlameVision}} \\
\cmidrule(lr){2-5}
 & \textbf{Accuracy(\%)} & \textbf{Precision(\%)} & \textbf{Recall(\%)} & \textbf{F1 score(\%)} \\
\midrule
$\alpha = 0.75, \beta = 1.0, \phi = 0.75, T=2.5$ & 97.33 & 97.62 & 97.33 & 97.39 \\
$\alpha = 0.3, \beta = 0.5, \phi = 0.3, T=1.5$ & 98.78 & 98.84 & 98.78 & 98.79 \\
$\alpha = 0.5, \beta = 0.7, \phi = 0.5, T=2$ & 94.78 & 95.18 & 94.78 & 94.88 \\
\bottomrule
\end{tabular}
\end{table*}

Table~\ref{tab:HumP-KD_param_settings} shows that the best performance is achieved with $\alpha=0.3, \beta=0.5, \phi=0.3, T=1.5$, achieving 98.78\% accuracy and 98.79\% F1 score. The configuration $\alpha=0.5, \beta=0.7, \phi=0.5, T=2$ results in poor performance with 94.78\% accuracy, highlighting the sensitivity of the proposed HumP-KD technique to hyperparameter selection. We defined empirically selected search ranges based on prior literature.

\begin{table*}[t]
\centering
\caption{Comparison of HumP-KD with existing fire classification methods on different datasets}
\label{tab:comparison_existing_work_merged}
\setlength{\tabcolsep}{6pt}
\renewcommand{\arraystretch}{1.3}
\begin{tabular}{c c c c p{5.8cm}}
\toprule
\textbf{Reference} & \textbf{Best Model} & \textbf{Accuracy (\%)} & \textbf{Other Metrics} & \textbf{Dataset} \\
\midrule
{\cite{sathishkumar2023forest}}  & Lwf-Xception & 91.41 & Not mentioned & Fire, WildfireSmoke, Fire and Smoke and BoW Fire Datasets \\
\midrule
{\cite{el2024real}} & DenseNet201 & 93.36 & F1 Score = 97.51\% & Wildfire  \\
\midrule
{\cite{ahmad2025firenet}} & FireNekt-KD & 95.16 & mAP@50 = 97.31\% & Flame  \\
\midrule
{\cite{titu2024real}}  & \shortstack{Knowledge Distillation-\\DET-R (YOLOv8n)} & 95.21 & F1 Score = 0.985 & Private Dataset \\
\midrule
{\cite{zia20243enb2}}  & 3ENB2 & 99.04 & Not mentioned & Deep-Quest-AI-2021, Carlo-2021, Bansal-2021 \\
\midrule
{\cite{11128176}}  & LAFNET & 55.7 & Recall = 51.5\% & FlameVision \\
\midrule
{\cite{10.1145/3744103.3744121}}  & FICNet & 78.4 & F1 Score = 78.4\% & FlameVision \\
\midrule
{\cite{jimaging12010043}}  & Mamba-YOLO & 89.6 & mAP@50 = 95.5\% & FlameVision \\
\midrule
\multirow{2}{*}{This work} 
& \multirow{2}{*}{\shortstack{HumP-KD \\ (MobileViT-S)}} 
& 94.78 & F1 Score = 94.88\% & FlameVision \\
\cmidrule(lr){3-5}
& & 98.48 & F1 Score = 98.48\% & Dataset-II \\
\bottomrule
\end{tabular}
\end{table*}

Table~\ref{tab:comparison_existing_work_merged} shows that HumP-KD achieves 94.78\% accuracy and 94.88\% F1 score on FlameVision, significantly outperforming prior methods such as LAFNET (55.7\%) and FICNet (78.4\%). HumP-KD attains 98.48\% accuracy on Dataset-II, demonstrating strong performance across different datasets.

To justify the selection of Swin teacher's $layer_{0,2}$ and ViT teacher's $block_6$, this work performed three analyses: \textbf{Linear Probing, Mutual Information Proxy, and Centered Kernel Alignment}. Linear probing evaluates the discriminative capability of feature representations by training a classifier head on frozen features. Mutual information proxy estimates the information content using the log-determinant of the covariance or Gram matrix. Centered kernel alignment measures similarity between feature representations using centered Gram matrices and normalized HSIC.

\begin{figure*}[t]
    \centering
    \includegraphics[width=1.0\textwidth]{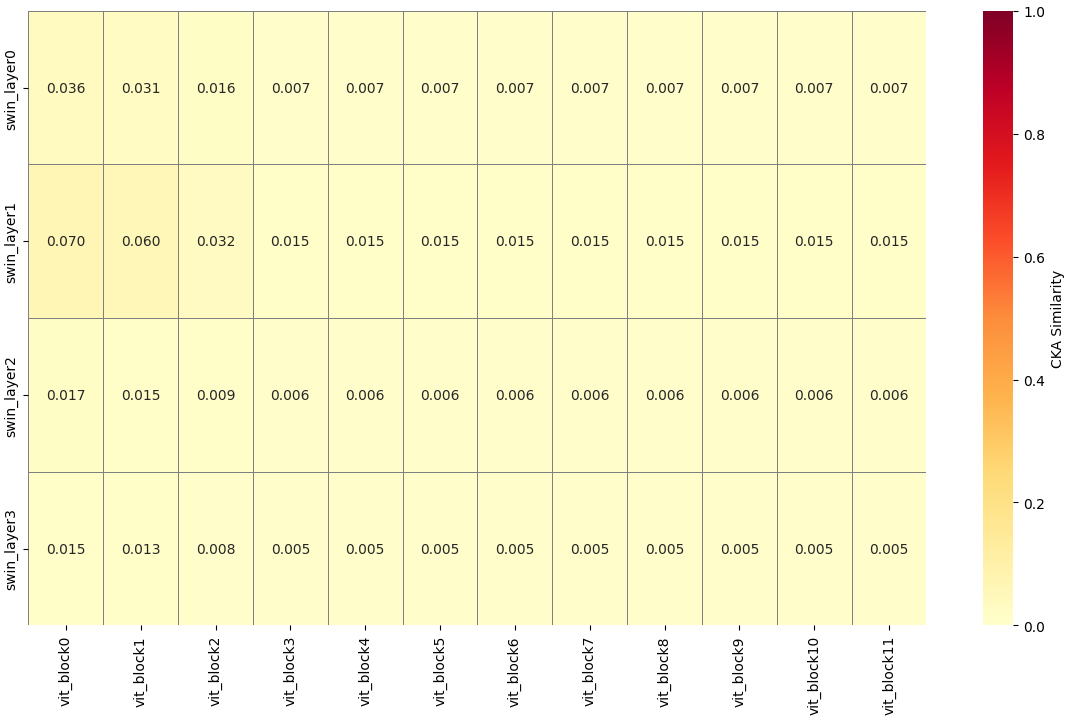}    
    \caption{Centered Kernel Alignment (CKA) similarity between Swin Transformer layers and ViT blocks.}
    \label{fig:swin_vit_cka}    
\end{figure*}

Figure~\ref{fig:swin_vit_cka} shows that the CKA similarity between Swin layers and ViT blocks is very low, ranging from approximately 0.005 to 0.070, indicating weak representational alignment.

\begin{figure*}[t]
    \centering
    \includegraphics[width=1.0\linewidth]{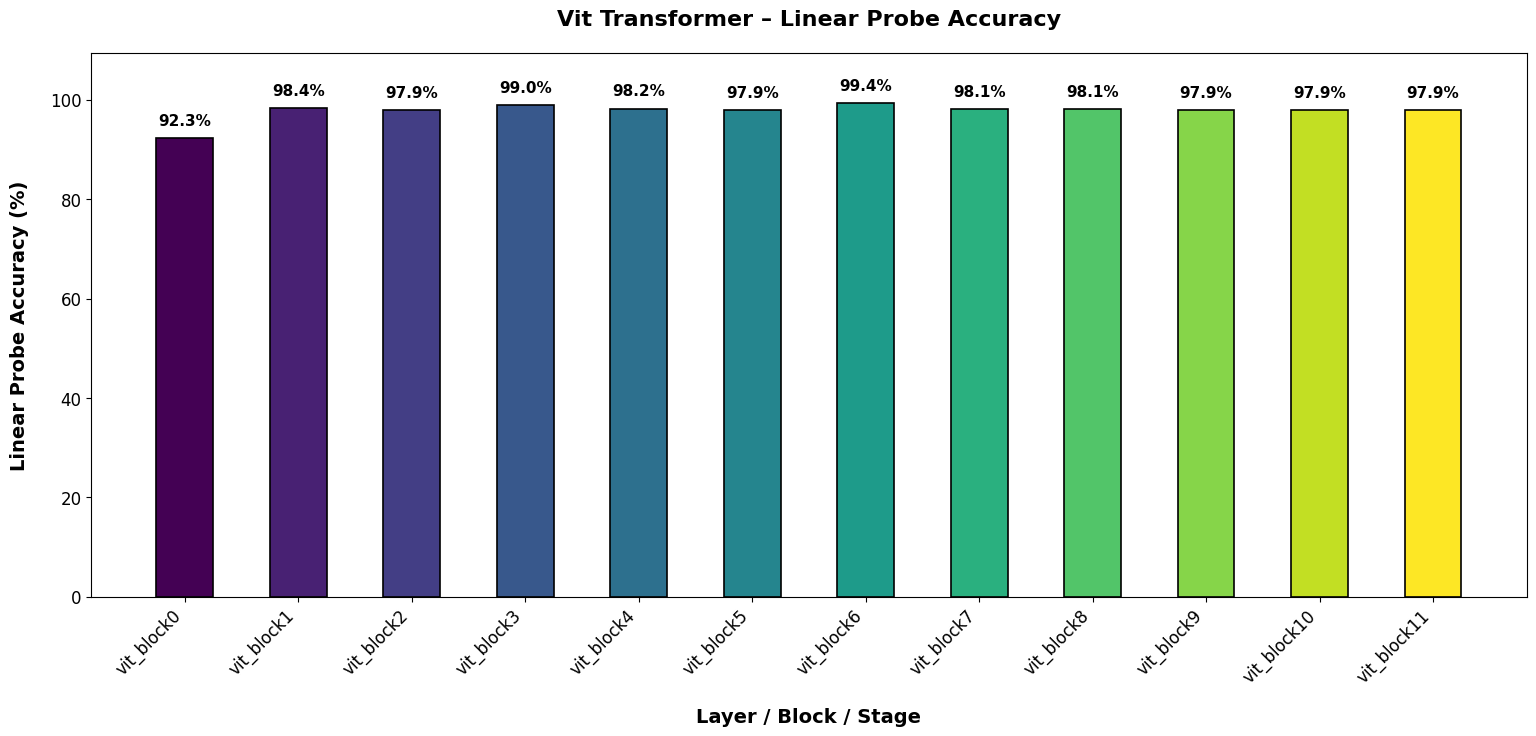}    
    \caption{Linear probing accuracy of ViT blocks showing discriminative representation across different layers.}
    \label{fig:vit_linear_probing}    
\end{figure*}

Figure~\ref{fig:vit_linear_probing} shows that the highest linear probing accuracy is achieved at block\_6 with 99.4\%, followed by block\_3 with 99.0\%. Earlier layers such as block\_0 achieve lower performance (92.3\%), indicating that deeper ViT representations are more discriminative..

\begin{figure*}[t]
    \centering
    \begin{minipage}{0.48\textwidth}
        \centering
        \includegraphics[width=\linewidth]{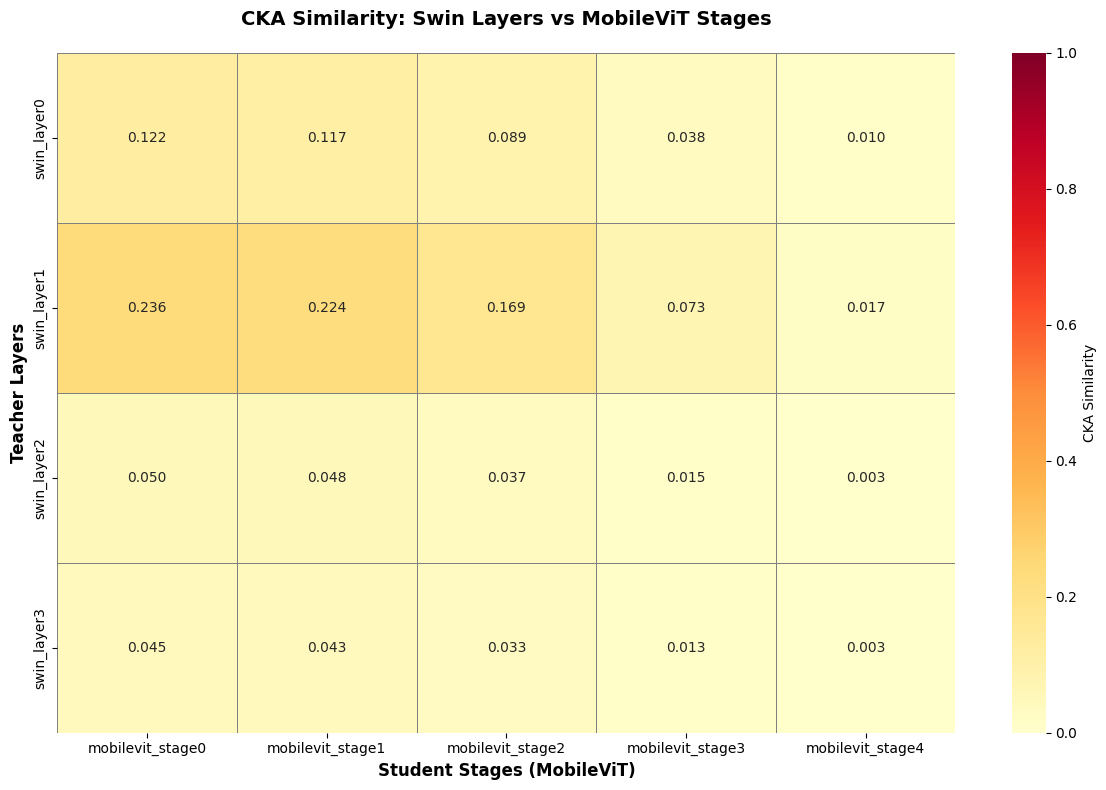}
        \centerline{(a) Swin-MobileViT CKA}
    \end{minipage}
    \hfill
    \begin{minipage}{0.48\textwidth}
        \centering
        \includegraphics[width=\linewidth]{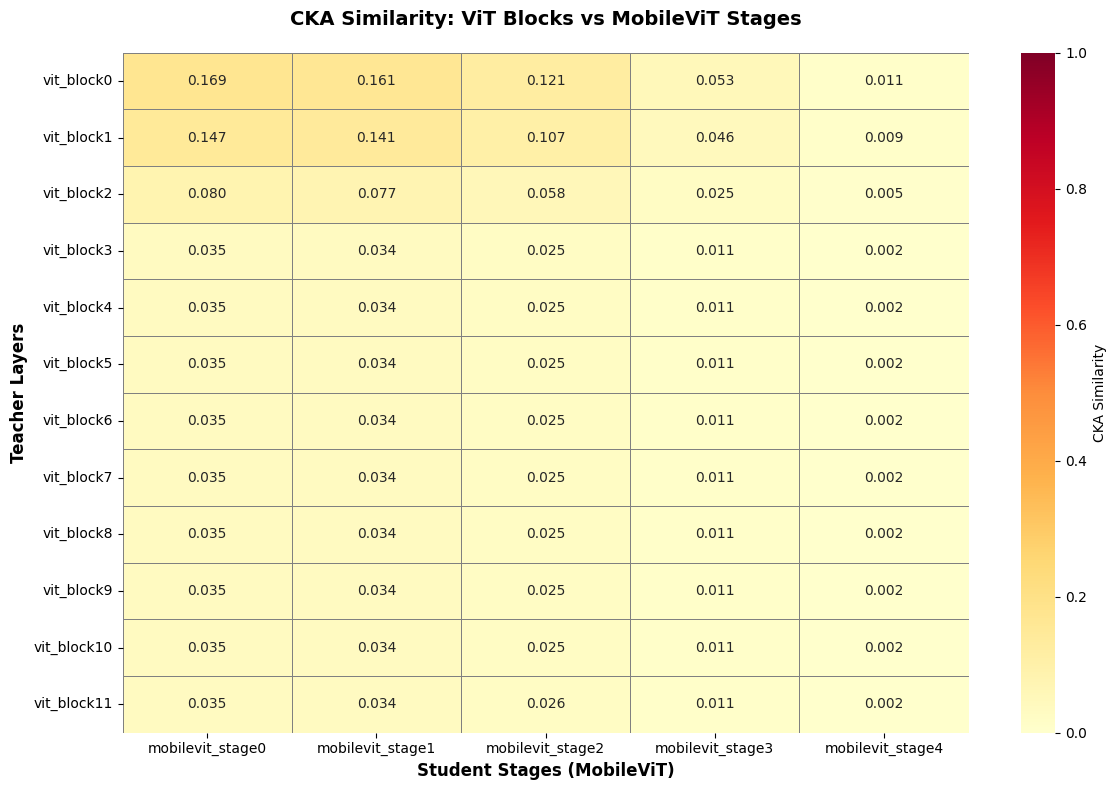}
        \centerline{(b)ViT-MobileViT CKA}
    \end{minipage}
    \caption{CKA similarity between teacher models (Swin and ViT) and the MobileViT student.}
    \label{fig:swin_mobilevit_vit_mobilevit_cka}    
\end{figure*}

Figure~\ref{fig:swin_mobilevit_vit_mobilevit_cka} shows that the highest CKA similarity is observed between Swin layer\_1 and MobileViT stage\_0 (0.236). ViT exhibits lower similarity values, decreasing from 0.169 to 0.002 across stages.

\subsection{T-Test/Wilcoxon}
The \textbf{T-Test} is a parametric test used to evaluate difference between the means of two groups.
The \textbf{Wilcoxon} signed-rank test is a non-parametric test used to evaluate differences between paired samples. 
This work applies both \textbf{T-Test} and \textbf{Wilcoxon} test on ten independent trials for both \textbf{HumP-KD} and \textbf{No-KD} (Table~\ref{tab:ttest_result}),(Table~\ref{tab:wilcoxon_result}), where No-KD refers to MobileViT trained using CrossEntropy. Using a threshold of $p < 0.05$, the results show that FlameVision is not statistically significant, while Dataset-II demonstrates a statistically significant difference.

\begin{table}[htb]
\centering
\caption{T-Test Result}
\label{tab:ttest_result}
\setlength{\tabcolsep}{3pt}
\renewcommand{\arraystretch}{1.4}
\begin{tabular}{l l l l}
\toprule
Metrics & & Flamevision & Dataset-II \\
\midrule
\multirow{2}{*}{Mean} & HumP-KD & $0.9637 \pm 0.0314$ & $0.9876 \pm 0.0063$ \\
 & No-KD & $0.8704 \pm 0.1634$ & $0.9537 \pm 0.0351$ \\
\midrule
\multirow{2}{*}{Median} & HumP-KD & 0.9633 & 0.9886 \\
 & No-KD & 0.9767 & 0.9697 \\
\midrule
t-stat & \multicolumn{2}{l}{1.7215} & 2.8364 \\
\midrule
p-value & \multicolumn{2}{l}{0.1193} & 0.0195 \\
\midrule
\shortstack[l]{Result \\ ( p $<$ 0.05)} & \multicolumn{2}{l}{Not Significant} & Significant \\
\bottomrule
\end{tabular}
\end{table}

\begin{table}[htb]
\centering
\caption{Statistical comparison of HumP-KD and No-KD using Wilcoxon signed-rank test across 10 trials}
\label{tab:wilcoxon_result}
\setlength{\tabcolsep}{3pt}
\renewcommand{\arraystretch}{1.4}
\begin{tabular}{l l l l}
\toprule
Metrics & & FlameVision & Dataset-II \\
\midrule
\multirow{2}{*}{Mean} & HumP-KD & $0.9637 \pm 0.0314$ & $0.9876 \pm 0.0063$ \\
 & No-KD & $0.8704 \pm 0.1634$ & $0.9537 \pm 0.0351$ \\
\midrule
\multirow{2}{*}{Median} & HumP-KD & 0.9633 & 0.9886 \\
 & No-KD & 0.9767 & 0.9697 \\
\midrule
W-stat & \multicolumn{2}{l}{13.0000} & 1.0000 \\
\midrule
p-value & \multicolumn{2}{l}{0.1602} & 0.0039 \\
\midrule
\shortstack[l]{Result \\ ( p $>$ 0.05)} & \multicolumn{2}{l}{Not Significant} & Significant \\
\bottomrule
\end{tabular}
\end{table}

Table~\ref{tab:ttest_result} and Table~\ref{tab:wilcoxon_result} shows that HumP-KD achieves higher and more stable performance, with $0.9876 \pm 0.0063$ accuracy on Dataset-II compared to $0.9537 \pm 0.0351$ for No-KD. The improvement is statistically significant on Dataset-II (for t-test $p=0.0195$, and for wilcoxon $p = 0.0039$), whereas the difference on FlameVision (for t-test $p=0.1193$, and for wilcoxon $p = 0.1602$) is not statistically significant.

\subsection{Attention Rollout}
\label{multi_scale_teacher_student_attention_rollout}

\begin{figure*}[!t]
    \centering
    \begin{minipage}{0.48\textwidth}
        \centering
        \includegraphics[width=\linewidth]{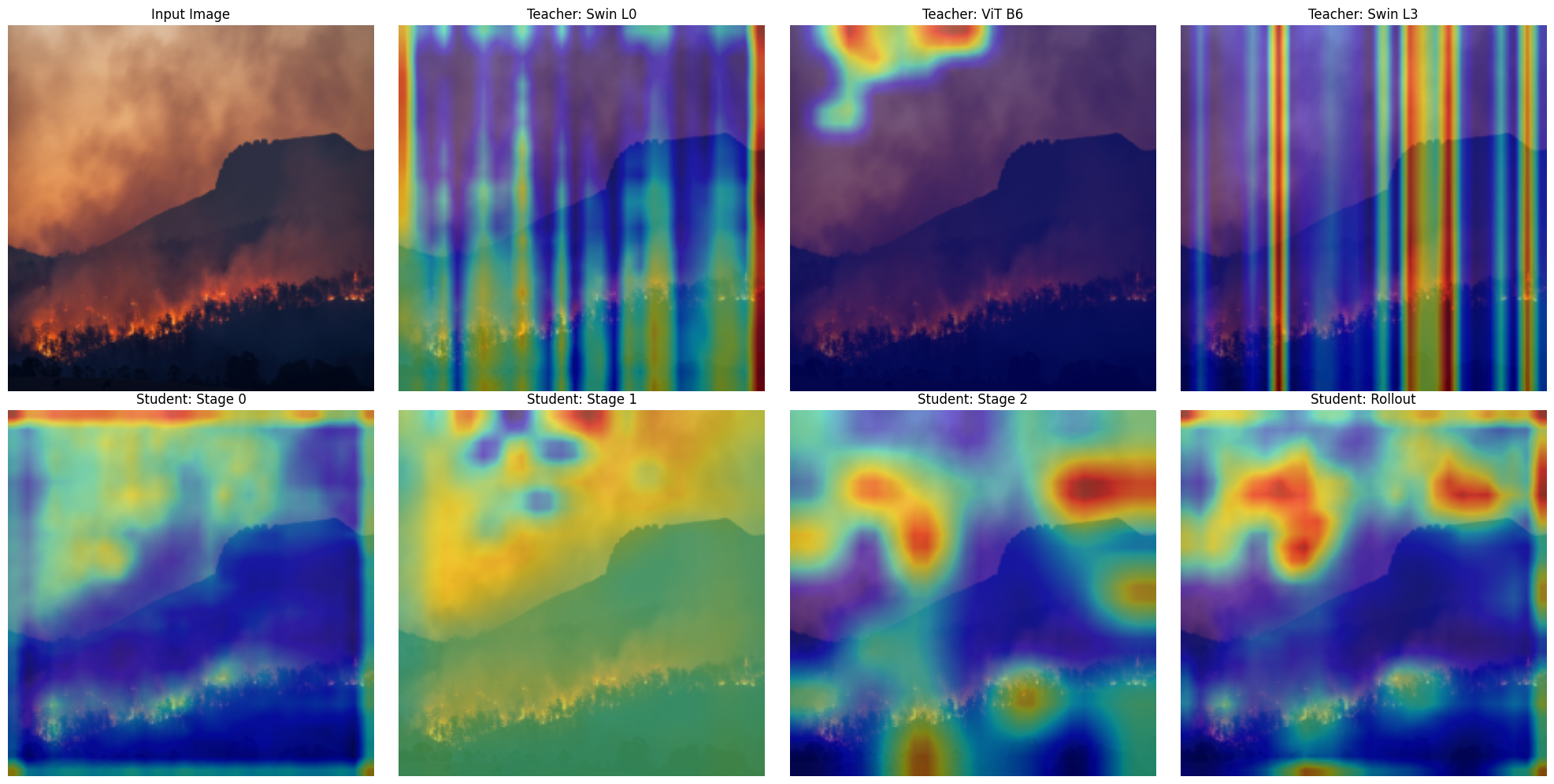}
        \centerline{(a) Fire}
    \end{minipage}
    \hfill
    \begin{minipage}{0.48\textwidth}
        \centering
        \includegraphics[width=\linewidth]{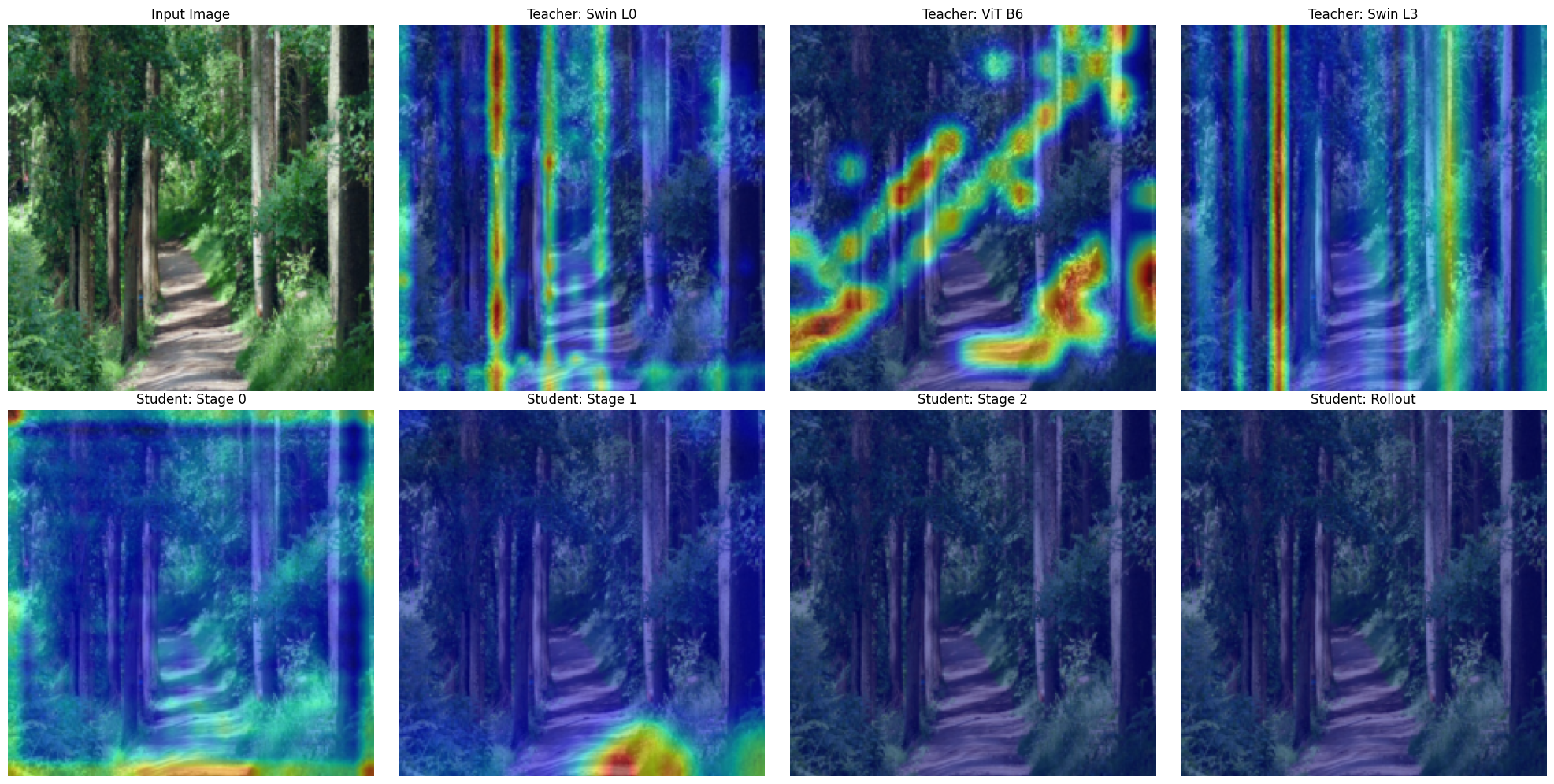}
        \centerline{(b) No-fire}
    \end{minipage}
    \caption{Multi-scale teacher–student attention rollout showing intermediate teacher layers and student stages for fire and no-fire inputs.}
    \label{fig:multiscale_teacher_studetn_attention_rollout}    
\end{figure*}

To synthesize a holistic saliency map for the student model, illustrated in Figure~\ref{fig:multiscale_teacher_studetn_attention_rollout}, we introduce a hierarchical feature interpretability framework to analyze the attention mechanisms of both teacher models (Swin Transformer and Vision Transformer) and the MobileViT student model. For each model, a Spatial Attention Map ($A$) is computed from intermediate feature maps $F \in \mathbb{R}^{B \times C \times H \times W}$ by aggregating channel-wise information using mean and ReLU:
\begin{equation}
A = \frac{\text{ReLU}\left(\frac{1}{C} \sum_{i=1}^{C} F_i\right)}{\sum_{h,w} \text{ReLU}\left(\frac{1}{C} \sum_{i=1}^{C} F_i\right) + \epsilon}
\end{equation}
For ViT, spatial reshaping is performed before attention aggregation. A hierarchical rollout mechanism is then applied by performing element-wise multiplication of attention maps across successive stages:
\begin{equation}
A_{rollout} = A_{stage2} \odot A_{stage1} \odot A_{stage0}
\end{equation}
The resulting map is normalized to highlight discriminative regions used by the student model. Attention maps from Swin-T ($L_0, L_3$) and ViT-B ($B_6$) are extracted and resized using bilinear interpolation to enable comparison with the student representation.

Figure~\ref{fig:multiscale_teacher_studetn_attention_rollout} presents normalized spatial attention maps from teacher models (Swin $layer_0$, ViT $block_6$, Swin $layer_3$) and the corresponding student stages. The student attention rollout is computed by progressively multiplying Stage 0–2 attention maps after spatial alignment, highlighting the regions emphasized by the HumP-KD model for (a) fire and (b) no-fire inputs.

\subsection{Real-Time Inference}
To verify the student model's performance beyond the training datasets, the HumP-KD student model is deployed in a web application and tested on unseen real-time images, including fire images, newspapers, park scenes, and flowers. Using Grad-CAM, the model generates heatmaps for both fire and no-fire images correctly. Since the model is trained primarily on images of devastating fires, it sometimes misclassifies images containing strong red, yellow, or orange colors. The model is deployed on a Flask API server in a web-based application named \textbf{FlameVision}, where users can upload an image and obtain the predicted label, confidence score, and highlighted regions.

\begin{figure}[htb]
    \centering
    \includegraphics[width=1.0\linewidth]{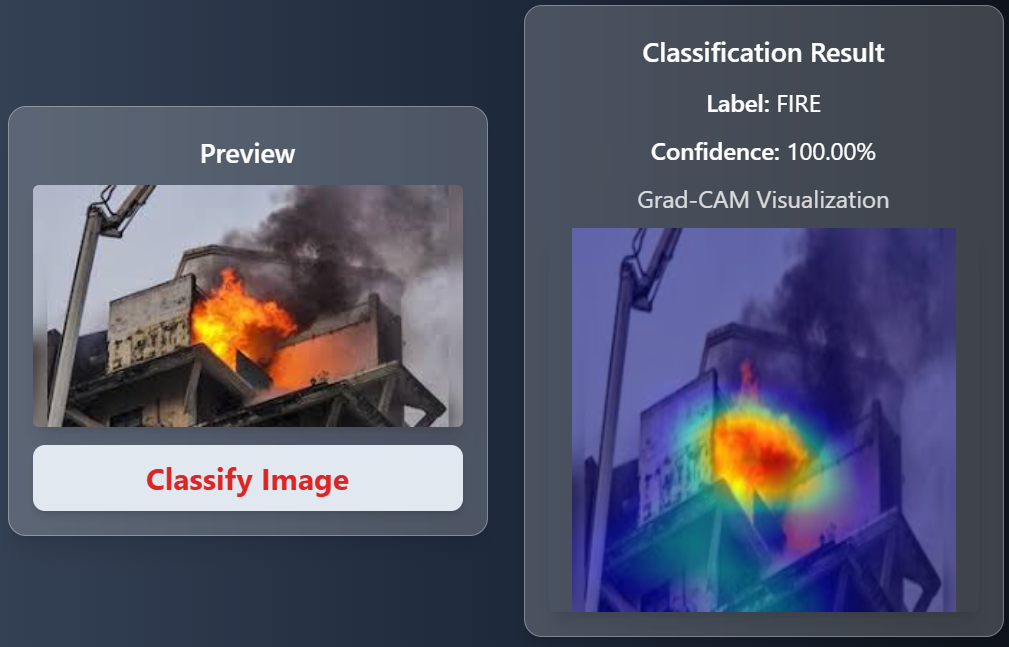}
    \caption{Web-based interface of the proposed FlameVision system demonstrating real-time fire detection with Grad-CAM visualization.}
    \label{fig:app_fire}   
\end{figure}

Figure~\ref{fig:app_fire} shows the web interface of the proposed FlameVision system, where the HumP-KD student model performs real-time fire classification. The model correctly predicts the input as fire with a confidence score of 100.00\% and highlights the discriminative fire region through Grad-CAM visualization.

\section{Conclusions}

Effective fire detection with lightweight models is essential to avoid fire-related accidents. In this work, we developed a Hybrid Uncertainty-aware Multi-stage Knowledge Distillation (HumP-KD) framework for real-time fire detection. We employed online augmentation techniques to enhance generality and applied several deep learning models, including CNN and Transformer-based architectures. From our evaluation results, the Swin Transformer performed the best when used individually. We combined Swin and ViT and utilized our HumP-KD pipeline to distill information into a lightweight student model, maximizing accuracy. Our research demonstrates that HumP-KD achieves a balance between efficiency and consistent performance, resulting in faster, safer fire classification systems for real-world use. Lastly, we deployed our model in the online application \textbf{FlameVision}, which enables users to upload fire images and utilize Grad-CAM to classify whether the image contains fire or not.

Our proposed system can be deployed into a mobile application to monitor real-time fire presence and avoid fire-causing accidents in advance. In the future, multiclass and multitask classification can be developed, which will open more ways to integrate our system into AI-based devices.

\bibliographystyle{IEEEtran}
\bibliography{ref}

@article{zia20243enb2,
  title={{3ENB2}: End-to-end {EfficientNetB2} model with online data augmentation for fire detection},
  author={Zia, Ehsanullah and Vahdat-Nejad, Hamed and Zeraatkar, Mohammad Ali and Joloudari, Javad Hassannataj and Hoseini, Seyyed Ali},
  journal={Signal, Image and Video Processing},
  volume={18},
  
  pages={7183--7197},
  year={2024},
  publisher={Springer}
}

@article{titu2024real,
  title={Real-time fire detection: Integrating lightweight deep learning models on drones with edge computing},
  author={Titu, Md Fahim Shahoriar and Pavel, Mahir Afser and Michael, Goh Kah Ong and Babar, Hisham and Aman, Umama and Khan, Riasat},
  journal={Drones},
  volume={8},
  
  pages={},
  year={2024},
  publisher={MDPI}
}

@Article{bioengineering11010070,
AUTHOR = {Zeng, Xinyi and Ji, Zhanlin and Zhang, Haiyang and Chen, Rui and Liao, Qinping and Wang, Jingkun and Lyu, Tao and Zhao, Li},
TITLE = {{DSP-KD}: Dual-Stage Progressive Knowledge Distillation for Skin Disease Classification},
JOURNAL = {Bioengineering},
VOLUME = {11},
YEAR = {2024}
}

@article{sathishkumar2023forest,
  title={Forest fire and smoke detection using deep learning-based learning without forgetting},
  author={Sathishkumar, Veerappampalayam Easwaramoorthy and Cho, Jaehyuk and Subramanian, Malliga and Naren, Obuli Sai},
  journal={Fire Ecology},
  volume={19},
  
  pages={},
  year={2023},
  publisher={Springer}
}

@ARTICLE{10839387,
  author={Choi, Sugi and Song, Youngjoo and Jung, Haiyoung},
  journal={IEEE Access}, 
  title={Study on Improving Detection Performance of Wildfire and Non-Fire Events Early Using Swin Transformer}, 
  year={2025},
  volume={13},
  number={},
  pages={46824-46837}
}

@article{yang2024real,
  title={Real-time fire and smoke detection with transfer learning based on cloud-edge collaborative architecture},
  author={Yang, Ming and Qian, Songrong and Wu, Xiaoqin},
  journal={IET Image Processing},
  volume={18},  
  pages={3716--3728},
  year={2024},
  publisher={Wiley Online Library}
}

@Article{fire8060211,
AUTHOR = {Ahmad, Naveed and Akbar, Mariam and Alkhammash, Eman H. and Jamjoom, Mona M.},
TITLE = {{CN2VF-Net}: A Hybrid Convolutional Neural Network and Vision Transformer Framework for Multi-Scale Fire Detection in Complex Environments},
JOURNAL = {Fire},
VOLUME = {8},
YEAR = {2025},
ISSN = {2571-6255}
}

@misc{flamevision2023,
  author       = {Anam Ibn Jafar and Al Mohimanul Islam and Fatiha Binta Masud and Jeath Rahmat Ullah and Md. Rayhan Ahmed},
  title        = {{FlameVision}: A new dataset for wildfire classification and detection using aerial imagery},
  howpublished = {Mendeley Data},
  year         = {2023},  
  doi          = {10.17632/fgvscdjsmt.4}
}

@INPROCEEDINGS{11128176,
  author={Zhai, Haozhou and Yan, Weiming and Wang, Xiaohan and Zhao, Tuhao and Hu, Tianjiang},
  booktitle={International Conference on Robotics and Automation}, 
  title={{LAFNET}: Lightweight Aerial Fire Detection Model for Onboard Edge Computing}, 
  year={2025},
  volume={},
  number={},
  pages={5150-5156}
}

@article{Gao2025YOLO11RLNAA,
  title={{YOLO11-RLN}: An aerial {UAV} algorithm for forest fire detection},
  author={Li Gao and Gaohua Chen},
  journal={Annals of the New York Academy of Sciences},
  year={2025}
}

@inproceedings{10.1145/3744103.3744121,
author = {Li, Shenzhi and Zhang, Biao and Zhao, Shuanglin and Wang, Peng},
title = {Research on Fire Classification and Detection Models Based on Deep Learning},
year = {2025},
booktitle = {International Symposium on AI and Cybersecurity},
pages = {85–90}
}

@article{el2024real,
  title={Real-time forest fire detection with lightweight {CNN} using hierarchical multi-task knowledge distillation},
  author={El-Madafri, Ismail and Pe{\~n}a, Marta and Olmedo-Torre, Noelia},
  journal={Fire},
  volume={7},  
  pages={},
  year={2024},
  publisher={MDPI}
}

@article{ahmad2025firenet,
  title={{FireNet-KD}: Swin Transformer-Based Wildfire Detection with Multi-Source Knowledge Distillation},
  author={Ahmad, Naveed and Akbar, Mariam and Alkhammash, Eman H and Jamjoom, Mona M},
  journal={Fire},
  volume={8},  
  pages={},
  year={2025},
  publisher={MDPI}
}

@INPROCEEDINGS{11071024,
  author={Li, Jiancheng and Lin, Wei},
  booktitle={International Conference on Computing, Networks and Internet of Things}, 
  title={Enhanced Fire Detection Using Knowledge Distillation and Multi-scale Attention in {YOLOv8}}, 
  year={2025},
  volume={},
  number={},
  pages={},
  }

@misc{minha2023forestfire,
  author       = {A. Minha},
  title        = {Forest Fire, Smoke and Non-Fire Image Dataset},
  howpublished = {Kaggle},
  year         = {2023},
  note         = {[Online; accessed 24 Feb 2025]},
  url          = {https://www.kaggle.com/datasets/amerzishminha/forest-fire-smoke-and-non-fire-image-dataset/data}
}

@article{el2023wildfire,
  title={The wildfire dataset: Enhancing deep learning-based forest fire detection with a diverse evolving open-source dataset focused on data representativeness and a novel multi-task learning approach},
  author={El-Madafri, Ismail and Pe{\~n}a, Marta and Olmedo-Torre, Noelia},
  journal={Forests},
  volume={14},  
  pages={},
  year={2023},
  publisher={MDPI}
}

@article{pavel2025multi,
  title={Multi-stage knowledge distillation with layer fusion-based deep learning approach for skin cancer classification},
  author={Pavel, Mahir Afser and Asad, Ramisa and Michael, Goh Kah Ong and Ikramuzzaman, Md and Mustakim, Murad and Khan, Riasat},
  journal={Scientific Reports},
  volume={15},  
  pages={},
  year={2025},
  publisher={Nature Publishing Group UK London}
}

@inproceedings{ousalah2025uncertainty,
  title={Uncertainty-aware knowledge distillation for compact and efficient 6dof pose estimation},
  author={Ousalah, Nassim Ali and Kacem, Anis and Ghorbel, Enjie and Koumandakis, Emmanuel and Aouada, Djamila},
  booktitle={International Conference on Intelligent Robots and Systems},
  pages={15398--15405},
  year={2025}
}

@Article{jimaging12010043,
AUTHOR = {Guo, Binhua and Liu, Dinghui and Shen, Zhou and Wang, Tiebin},
TITLE = {{FF-Mamba-YOLO}: An {SSM}-Based Benchmark for Forest Fire Detection in {UAV} Remote Sensing Images},
JOURNAL = {Journal of Imaging},
VOLUME = {12},
YEAR = {2026}
}

@ARTICLE{10795207,
  author={Yang, Yang and Wang, Chao and Gong, Lei and Wu, Min and Chen, Zhenghua and Gao, Yingxue and Wang, Teng and Zhou, Xuehai},
  journal={IEEE Transactions on Circuits and Systems for Video Technology}, 
  title={Uncertainty-Aware Self-Knowledge Distillation}, 
  year={2025},
  volume={35},  
  pages={4464-4478}
  }

@Article{app12199453,
AUTHOR = {Xu, Chuanyun and Bai, Nanlan and Gao, Wenjian and Li, Tian and Li, Mengwei and Li, Gang and Zhang, Yang},
TITLE = {Multiple-Stage Knowledge Distillation},
JOURNAL = {Applied Sciences},
VOLUME = {12},
YEAR = {2022}
}

@misc{dailyStarFire2025,
  author    = {{The Daily Star}},
  title     = {Bangladesh saw 75 fires a day in 2025: Fire Service},
  year      = {2025},
  howpublished = {\url{https://www.thedailystar.net/news/bangladesh/accidents-fires/news/bangladesh-saw-75-fires-day-2025-fire-service-4102796}},
  note      = {Accessed: 2026-04-03}
}
\end{document}